\pdfoutput=1
\documentclass[11pt]{article}

\usepackage{ACL2024} %[review]

\usepackage{times}
\usepackage{latexsym}
\usepackage{threeparttable}

\usepackage[T1]{fontenc}
\usepackage[utf8]{inputenc}

\usepackage{microtype}

\usepackage{inconsolata}

\usepackage{soul}
\usepackage{adjustbox}
\usepackage{tabularx}
\usepackage{multirow}
\usepackage{multicol}
\usepackage{booktabs}
\usepackage{array}
\usepackage{makecell}
\usepackage{amsmath,amsfonts,mathtools}

\usepackage{algpseudocodex, algorithm, float}

\makeatletter
\newcommand*{\Xbar}{}%
\DeclareRobustCommand*{\Xbar}{%
  \mathpalette\@Xbar{}%
}
\newcommand*{\@Xbar}[2]{%
  \sbox0{$#1\mathrm{X}\m@th$}%
  \sbox2{$#1X\m@th$}%
  \rlap{%
    \hbox to\wd2{%
      \hfill
      $\overline{%
        \vrule width 0pt height\ht0 %
        \kern\wd0 %
      }$%
    }%
  }%
  \copy2 %
}
\makeatother

\title{{An Investigation of Neuron Activation as a Unified Lens to Explain Chain-of-Thought Eliciting Arithmetic Reasoning of LLMs}}

\author{Daking Rai \\
  Department of Computer Science \\
  George Mason University \\
  Fairfax, VA \\
  \texttt{drai2@gmu.edu} \\\And
  Ziyu Yao \\
  Department of Computer Science \\
  George Mason University \\
  Fairfax, VA \\
  \texttt{ziyuyao@gmu.edu} \\}

\begin{document}
\maketitle
\begin{abstract}

Large language models (LLMs) have shown strong arithmetic reasoning capabilities when prompted with Chain-of-Thought (CoT) prompts. However, we have only a limited understanding of how they are processed by LLMs. To demystify it, prior work has primarily focused on ablating different components in the CoT prompt and empirically observing their resulting LLM performance change \cite{madaan2022text, wang-etal-2023-towards, ye-etal-2023-complementary}. Yet, the reason why these components are important to LLM reasoning is not explored. To fill this gap, in this work, we investigate ``neuron activation'' as a lens to provide a unified explanation to observations made by prior work. Specifically, we look into neurons within the feed-forward layers of LLMs that may have activated their arithmetic reasoning capabilities, using Llama2 \cite{touvron2023llama} as an example. To facilitate this investigation, we also propose an approach based on GPT-4 to automatically identify neurons that imply arithmetic reasoning. Our analyses revealed that the activation of reasoning neurons in the feed-forward layers of an LLM can explain the importance of various components in a CoT prompt, and future research can extend it for a more complete understanding.\footnote{The source code for our implementation is available at \url{https://github.com/Dakingrai/neuron-analysis-cot-arithmetic-reasoning}.} 

\end{abstract}

\section{Introduction}
Arithmetic reasoning is one of the emergent properties in large language models (LLMs), which is necessary for them to tackle tasks that require multiple steps to arrive at the correct answer. In recent years, Chain-of-Thought (CoT) has become a popular prompting strategy to elicit reasoning\footnote{Our work focuses on ``arithmetic reasoning''. For ease of presentation, we use ``reasoning'' interchangeably with it.} in LLMs \cite{wei2022chain}. Despite its successes, there is little understanding of what makes it effective and how LLMs utilize it to facilitate reasoning. 

To address this concern, a line of research has focused on decomposing the CoT prompt into various components and performing ablation studies on them to ascertain the significance of each component on the LLM reasoning performance \cite{madaan2022text, wang-etal-2023-towards, ye-etal-2023-complementary}.
% analyzing the input and output of CoT prompts by decomposing the prompt into different components and doing an ablation study to ascertain their significance on LLM performance 
Although these studies have yielded several insightful observations on the effect of input on LLM's reasoning performance, they do not shed light on how these inputs are being processed internally by LLMs to perform reasoning. 

On the other hand, there is a growing body of research in the field of mechanistic interpretability \cite{elhage2021mathematical, wang2022interpretability} that specifically examines the internals of LLMs to understand their mechanism. In this vein, \citet{stolfo2023mechanistic} studied the internal mechanism of LLMs to perform arithmetic calculation, suggesting that attention heads facilitate information traversal, while the feed-forward layer (FF) handles information processing to produce accurate answers for a given computation. However, \citet{stolfo2023mechanistic} only studied the mechanism for a single mathematical computation and doesn't study arithmetic reasoning in full scope. In parallel, some other research demonstrated that LLMs consist of neurons that can be associated with human-interpretable concepts, which play a crucial role in various capabilities of LLMs \cite{geva2022transformer, dai2021knowledge, gurnee2024universal}. Specifically, \citet{geva2022transformer} showed that neurons in the FF layer of a transformer model~\cite{vaswani2017attention} form key-value pairs that facilitate next-token prediction by promoting concepts in the vocabulary space. 
% They then successfully applied this intuition to suppress the toxic language generation of LLMs by manually activating the neurons that promote safe words during inference time. 
However, none of the prior work has applied the intuition to understand LLM reasoning.

Motivating by the need to form a deeper understanding of how CoT prompts elicit reasoning in LLMs and observing the pivotal role of neurons within the FF layers of LLMs \cite{geva2022transformer}, in this work, we propose to investigate the activation of FF neurons in LLMs as a lens to interpret their arithmetic reasoning capabilities. Particularly, we aim to use neuron activation to provide a unified explanation of observations that were only empirically made by prior work~\cite{madaan2022text, wang-etal-2023-towards, ye-etal-2023-complementary}, as listed in Table~\ref{tab:prior_research}. 

To this end, we first propose an approach that leverages LLMs (e.g., GPT-4~\cite{openai2023gpt}) to \emph{automatically} search for neurons that are related to arithmetic reasoning (e.g., arithmetic addition, logical connections, etc.). Prior work trying to search for concept-relevant neurons relies on human analysis \cite{geva2022transformer, elhage2022solu}. For example, \citet{geva2022transformer} proposed to manually examine a neuron's top promoted tokens and determine if the neuron promotes the given pre-defined concept or not. However, this manual approach becomes impractical for LLMs with a large number of layers and numerous neurons in each layer. Our approach instead decides whether a given neuron expresses a certain concept automatically by prompting the GPT-4 with its top promoted tokens and make a judgment on its represented concept. Our experimental results demonstrate the high effectiveness of utilizing GPT-4 for this purpose. Subsequently, we apply our proposed approach to identify FF neurons in Llama2-7B that promote several concepts relevant to arithmetic reasoning, listed in Table~\ref{tab:Neurons}.

Leveraging the identified reasoning neurons, we performed a series of analyses on observations made by prior work~\cite{madaan2022text, wang-etal-2023-towards, ye-etal-2023-complementary}, including the importance of textual explanation, equations, arithmetic diversity, and the negligible impact of incorrect labels in CoT prompts for elicitating reasoning in LLMs. Specifically, we analyzed the activation patterns of the identified reasoning neurons, such as their activation frequency and strength, to gain insights into these observations. Our results reveal that examining the activation of FF neurons in response to different CoT prompts can provide valuable insights into why certain CoT prompts are more effective in eliciting arithmetic reasoning capabilities in LLMs. We then conclude the paper with a discussion of future work that can complement the proposed neuron activation analysis with other approaches to form a more complete understanding of LLM reasoning.

\section{Background and Related Work}
% \subsection{Why neuron analysis?}
% Prior studies attempt to understand the reasoning in LLMs by decomposing the Chain-of-thought (CoT) prompt into different semantic components and evaluating their importance for eliciting reasoning in LLMs. For instance, \citet{ye-etal-2023-complementary} decomposes the GSM8K \cite{cobbe2021training} examples into equations and text, whereas \citet{madaan2022text} decomposes them into symbols, equations, and text. The importance of these components is evaluated by conducting ablation studies or generating counterfactual examples where a higher impact on the model's performance indicates a higher component's importance in aiding reasoning in LLMs. We present the summary of previous studies' findings in Table \ref{tab:prior_research}. 

\subsection{Prior Work towards Understanding the CoT Reasoning of LLMs}\label{subsec:related-work}

\begin{table*}[ht]
    \centering
    \begin{threeparttable}
    \resizebox{\textwidth}{!}{%
    \begin{tabular}{>{\centering\arraybackslash}p{3.5cm}p{13cm}>{\centering\arraybackslash}p{3cm}>{\centering\arraybackslash}p{1.5cm}}
    \toprule
    \textbf{Research Questions} & \textbf{Examples in CoT Prompts} & \textbf{Prior Work} & \textbf{Findings}  \\
    \toprule

    \textit{Does equation matter? (RQ3)} &
    {
    \textbf{w Equation:} Let's think step by step. First there are 15 trees. Then there were 21 trees after some more were planted. So there must have been \textbf{21 - 15 = 6} trees. The answer is 6. \newline
    \textbf{w/o Equation:} Let's think step by step. First there are 15 trees. Then there were 21 trees after some more were planted. So there must have been \textbf{6} trees. The answer is 6.}
    & \citet{wang-etal-2023-towards, ye-etal-2023-complementary, madaan2022text} & Yes \\
     \midrule
     
    \textit{Does textual explanation matter? (RQ4)} &
    {
    \textbf{w Textual Explanation:} Let's think step by step. First Leah had 32 chocolates and her sister had 42 chocolates. So in total they had 32 + 42 = 74 chocolates. Then they ate 35 chocolates. So there must be 74 - 35 = 39 chocolates. The answer is 39. \newline
    \textbf{w/o Textual Explanation:} 32 + 42 = 74. 74 - 35 = 39. The answer is 39.}
    & \citet{wang-etal-2023-towards, ye-etal-2023-complementary, madaan2022text} & Yes\\
    
    \midrule

    \textit{Does the diversity of arithmetic operators matter? (RQ5)} & 
    {\textbf{AddOnly:} Let's think step by step. First there are 3 cars. Then 2 more cars arrive. So there must be 3 + 2 = 5 cars. The answer is 5. \newline
    \textbf{MultOnly:} Let's think step by step. First a farmer has 5 cows. Then each cow has 4 legs. So the cows have 5 x 4 = 20 legs in total. The answer is 20.}
    & \citet{ye-etal-2023-complementary} & Yes \\ 
    
    \midrule
    \textit{Does incorrect reasoning or gold label not matter? (RQ6)} & 
     {\textbf{Correct Label:} Let's think step by step. First there are 15 trees. Then there were 21 trees after some more were planted. So there must have been 21 - 15 = 6 trees. The answer is 6. \newline
     \textbf{Incorrect Label:} Let's think step by step. First there are 15 trees. Then there were 21 trees after some more were planted. So there must have been 21 - 15 = \textbf{1} trees. The answer is \textbf{1}.}
     & \citet{wang-etal-2023-towards, ye-etal-2023-complementary} & No
     \\

     \cmidrule{2-4}
     & 
     {
     \textbf{OOD Label:} Let's think step by step. First there are 15 trees. Then there were 21 trees after some more were planted. So there must have been 21 - 15 = \textbf{Dawson} trees. The answer is \textbf{Dawson}.} 
     & \citet{wang-etal-2023-towards, ye-etal-2023-complementary} & Yes \\

    \bottomrule
    \end{tabular}
    }
    \caption{Summary of shared findings from prior works. Our reproduced results are shown in Table~\ref{tab: accuracy}.}
    \label{tab:prior_research}
\end{threeparttable}
\end{table*}

Prior studies attempted to understand the arithmetic reasoning in LLMs by decomposing the Chain-of-Thought (CoT) prompt into different semantic components and evaluating their importance via ablation studies. We present a summary of the major findings from prior work in Table~\ref{tab:prior_research}. For example, to understand whether equations matter in the few-shot CoT prompt, \citet{ye-etal-2023-complementary} experimented with a CoT variant where all equations (e.g., ``21 - 15 = 6'') were eliminated and only the calculation results (e.g., ``6'') was presented. By observing the resulting LLM performance change, one can empirically gauge the importance of equations in a CoT prompt. While previous studies have highlighted the significance of various components (e.g., textual explanation, equations, etc.) within the CoT prompt, the underlying reason behind these observations remains unanswered. This thus motivates us to study the underlying inner mechanism that is responsible for LLM reasoning.

\subsection{Interpreting Neurons of LLMs}
Many prior interpretability works have studied neurons to understand the inner mechanism of LLMs and have led to the discovery of many interesting types of neurons such as knowledge neurons \cite{dai2021knowledge}, skill neurons \cite{wang2022finding}, sentiment neurons \cite{radford2017learning}, concept neurons \cite{geva2022transformer}, universal neurons \cite{gurnee2024universal}, and many others related to linguistic and grammar features \cite{durrani2022linguistic, sajjad2022analyzing}. Furthermore, the activation patterns of these neurons have been found to significantly influence the behavior of LLMs \cite{geva2022transformer}. 
To discover the targeted neurons, probing is the most widely used approach, which involves training a simple classifier (probe) on the representations of neurons using a human-annotated dataset \cite{gurnee2023finding, belinkov2022probing}. Another popular approach specific to transformer-based LLMs is the projection of neuron representations to the vocabulary space, introduced by \citet{geva2022transformer}, and has been widely adopted \cite{dar2022analyzing, belrose2023eliciting, ghandeharioun2024patchscope}.
 
However, to the best of our knowledge, none of the prior work has applied neuron activation to understand LLM reasoning. Our work draws inspiration from~\citet{geva2022transformer} but extends it for a unified explanation of observations in CoT prompting. To this end, we also proposed an automatic approach based on GPT-4 for neuron discovery.

Relevant to our work, \citet{stolfo2023mechanistic} have also attempted to understand arithmetic reasoning by interpreting their neuron behaviors. However, the majority of their study focused on 
% higher-level analysis units including attention blocks and MLP blocks. 
coarser units such as the entire attention or FF layer. Furthermore, their investigation solely focused on how LLMs execute arithmetic calculations whereas the (multi-step) reasoning process is under-explored.

\subsection{Concept Promotion via Neuron Activation of \citet{geva2022transformer}}
\label{sec:sec2.2}
Our work builds upon the findings of \citet{geva2022transformer}, which shows the role of the feed-forward (FF) layer in the construction of an LLM's prediction -- (a) each FF layer induces \emph{additive updates} to token representations, which can be further decomposed into weighted collections of \emph{sub-updates}; (b) both the token representation and sub-updates of the FF layer can be projected at any stage to a distribution over the output vocabulary. Through the vocabulary space projection, the authors found that the sub-updates of an FF layer often encode human-interpretable concepts.
% Furthermore, their study also shows that these sub-updates often encode human-interpretable concepts after we project them in vocabulary space which makes it a useful technique for understanding LLMs' behavior. 
% Next, we formally define \citet{geva2022transformer}'s findings.
Next, we briefly describe \citet{geva2022transformer}'s findings; more details should refer to the original paper.

% Consider a given auto-regressive transformer-based LLM $G: X \rightarrow Y$ over vocabulary $V$ that consists of $L= l_1,...l_n, l_i \in L$ layers and embedding matrix $E \in \mathbb{R}^{d \times |V|}$ of hidden dimension $d$, over a vocabulary $V$. $G$ maps a token sequence $X=x_1,...x_t$, $x_i \in V$ to a probability distribution $y \in Y \subset \mathbb{R}^{|V|} \rightarrow [0,1] $ that predicts next token given the input sequence $x$. Then the hidden representation $x_i^l$ of the token $x_i$ at each layer $l$ is processed by $FFN^l$ and subsequently added it's output to the hidden representation:
Consider an auto-regressive transformer-based LLM consisting of $L$ layers, which predicts the next token by projecting its last-layer hidden state onto a vocabulary $\mathcal{V}$ via an embedding matrix $E \in \mathbb{R}^{d \times |\mathcal{V}|}$, where $d$ denotes the embedding size and $|\mathcal{V}|$ represents the vocabulary size. We denote the FF component in the $l$-th layer as $FF^l$. Given a token sequence $X = (x_1, ..., x_{|X|})$ as input, the representation of each token $x_i$ at layer~$l$ (denoted as $x_i^l \in \mathbb{R}^d$) is updated by $FF^l$ as follows:
\vspace{-2mm}
\begin{equation}
\small
\bar{x}_i^l = x_i^l + FF^l(x_i^l)
\label{eq:eq1}
\vspace{-2mm}
\end{equation}
The updated representation $\bar{x}_i^l$ then goes through the multi-head self-attention at layer $l$, which results in $x_i^{l+1}$ for the next FF layer (i.e., $FF^{l+1}$). With the residual connection~\cite{he2016deep}, each FF update can be seen as producing \emph{additive updates} to the token representation.
% where $x_i^l$ is the residual stream after the preceding multi-head self-attention layer while $\bar{x}^l_i$ is the residual stream after the additive update to token representation by $FFN^l$. Each $FFN^l$ is defined with two parameter matrices $K^l, V^l \in \mathbb{R}^{d_m \times d}$ where $d_m$ is the intermediate hidden dimension and non-linearity function $f$:

In transformers, each $FF^l$ is defined with two parameter matrices $K^l, V^l \in \mathbb{R}^{d_m \times d}$, where $d_m$ is the intermediate hidden dimension, and a non-linearity function $f$: 
\vspace{-2mm}
\begin{equation}\small
FF^l(x_i^l) = f(K^lx_i^l)V^l
\label{eq:eq2}
\vspace{-2mm}
\end{equation}
% \citet{geva2022transformer} interpreted $FFN^l$ output by decomposing it into a set of $d_m$ \emph{sub-updates}, each induced by a single FFN parameter vector. Formally, 
Eqn~\ref{eq:eq2} can further be decomposed as:
\vspace{-2mm}
\begin{equation}\small
    FF^l (x_i^l)= \sum_{j=1}^{d_m} f(x_i^l \cdot k^l_j) v^l_j = \sum_{j=1}^{d_m} m^l_{ij} v^l_j
    \label{eq:eq3}
\vspace{-2mm}
\end{equation}
where $k_j^l \in \mathbb{R}^d$ and $v_j^l \in \mathbb{R}^d$ are the $j$-th row of $K^l$ and $V^l$, respectively, and $m^l_{ij}=f(x_i^l \cdot k_j^l)$ is a scalar representing the \emph{activation coefficient} of $v^l_j$ (i.e., the \emph{neuron}). \citet{geva2022transformer} interpreted each term in this sum as a set of $d_m$ \emph{sub-updates} to the token representation. They also proposed to project this sub-update to the vocabulary by $Ev^l_j$. By analyzing the projected vocabulary tokens (typically tokens with top projection scores), they found that the sub-update often encodes human-interpretable concepts. It is important to note that every $v^l_j$ is a static parameter that is input-independent, while the coefficient $m^l_{ij}$ depends on the input token $x_i$. 
% while sub-updates are dynamic as they are weighted by input-dependent coefficients.

Observing their critical roles and leveraging their interpretability after projection, \citet{geva2022transformer} demonstrated the potential of encouraging non-toxic language by manipulating the coefficients of FF neurons in LLMs. This was achieved by identifying FF neurons representing non-toxic language concepts and then increasing their coefficients. Getting inspired by their findings, our work aims to explore: Can FF neuron activation be similarly used to interpret and even control LLM reasoning? It is important to note that ``toxicity'' and ``reasoning'' represent distinct extents of abstraction. While whether a sentence is toxic or not can be judged by superficial keyword searching, ``reasoning'' is more abstract and can encompass multiple aspects (e.g., logical induction, mathematical calculation, etc.), which thus presents a significant challenge.

\section{Neuron Discovery using GPT-4} \label{sec: neuron-discovery-algo}
To facilitate the neuron analysis, we first propose an approach for discovering neurons that express concepts related to arithmetic reasoning.
To achieve the same goal, \citet{geva2022transformer} manually examined the top-scoring vocabulary tokens projected by each neuron $v^l_j$ and annotated its concept. However, this manual search approach can become impractical for LLMs with deep layers and numerous sub-updates per layer.
To overcome this inefficiency, we propose a method that leverages GPT-4 to automate the search process.

Our proposed approach involves two steps. First, for a given LLM, we store the $T$ neurons $v^l_j$'s with the largest coefficient $m^l_{ij}$ from each layer $l$ and at each generation time step $i$, using a set of examples $\mathcal{E}$ that showcase the LLM's capability (i.e., arithmetic reasoning in our case) to provide the prompt. We only considered the top-$T$ neurons to narrow our search to the most activated neurons. This returns a set of candidate neurons $\mathcal{N}$. We present this step in Algorithm~\ref{alg: one}.

\begin{algorithm}[t!]
    \caption{Candidate Neuron Collection}\label{alg: one}
    \begin{small}
    \begin{algorithmic}[1]
            \State \textbf{Input:} A set of examples $\mathcal{E}$ implying the capability, a filtering threshold $T$, the target LLM
            \State \textbf{Output:} A set of candidate neurons $\mathcal{N}$.
            \State Initialize $\mathcal{N} \leftarrow \{\}$
            \For{{each example in $\mathcal{E}$}}:
                \For{{each decoding step $i$}}:
                    \For{{each layer $l=1,...,L$}}:
                        \State $\{m^l_{ij^\prime}\}_{j^\prime=1}^T$ \hspace{-2mm}  $\leftarrow$ $\texttt{FindLargestT}(\{m^l_{ij}\}_{j=1}^{d_m}, T)$
                        \State $\mathcal{N} \leftarrow  \mathcal{N}  \cup \{v^l_j | m^l_{ij} \in \{m^l_{ij^\prime}\}_{j^\prime=1}^T \}$
                    \EndFor
                \EndFor
            \EndFor
    \end{algorithmic}
    \end{small}
\end{algorithm}

\begin{algorithm}[t!]
    \caption{Neuron Annotation via GPT-4}\label{alg: two}
    \begin{small}
    \begin{algorithmic}[1]
            \State \textbf{Input:} Concept $C_{name}$, a set of seed tokens $\mathcal{S}_{name}$, filtering thresholds $P$ and $F$, embedding $E$ of LLM, and candidate neuron set $\mathcal{N}$. 
            \State \textbf{Output:} A subset of neurons $\mathcal{R} \subset \mathcal{N}$ representing concept $C_{name}$.
            \State Initialize $\mathcal{R} \leftarrow \{\}$
            \For{{each neuron $v_n \in \mathcal{N}$}}:
                % \State Save the top-$p$ token promotions of $m_i$ after projecting $v_i$ to the vocabulary space, as described in Section \ref{tab:prior_research}, denoted as $T=\{t_1, .., t_{p}\}$.
                \State $\mathcal{V}_P=\{w_1,...,w_P\} \leftarrow \texttt{GetLargestP}(Ev_n, P)$
                \If {$|\mathcal{V}_P \cap \mathcal{S}_{name}| \geq F$}:
                    \If {$\texttt{GPT4ConceptQuery}(\mathcal{V}_P, C_{name})$}:
                    % \State Query GPT-4 with the following prompt: 
                    % \textit{``A neuron in language model promotes the following set of words: $t_1, .., t_{p}$. Is this neuron promoting $C_{concept}$? First, answer in Yes or No format and provide an explanation."}
                    % \State If the GPT-4 response is ``Yes", then $R \leftarrow R \cup \{n_i\}$. Else, Continue.
                        \State $\mathcal{R} \leftarrow \mathcal{R} \cup \{v_n\} $
                    \EndIf
                \EndIf
            \EndFor
    \end{algorithmic}
    \end{small}
\end{algorithm}

In the second step, we task GPT-4 to determine whether each neuron in $\mathcal{N}$ promotes a predefined concept $C_{name}$ (e.g., arithmetic addition). However, employing GPT-4 to classify all neurons in $\mathcal{N}$ still requires a large number of prompts and may incur significant costs. To address this issue, we propose to first filter out the irrelevant neurons by using a set of human-annotated ``seed tokens'' (denoted as $\mathcal{S}_{name}$) that are likely to be associated with the given concept as per human intuition.\footnote{{Seed tokens are solely utilized to filter out irrelevant neurons. They are not included in the prompt for GPT-4 and do not influence the neuron annotation process.}} For instance, when searching for neurons that promote arithmetic addition, relevant tokens may include ``add'', ``addition'', ``sum'', ``+'', and ``plus''. Although a neuron that promotes the given concept may not invariably promote all the tokens from the seed tokens, it is quite probable that it promotes at least some of them. Leveraging this insight, we filter out neurons that do not consist of at least a threshold of $F$ seed tokens in their top-$P$ promoted tokens $\mathcal{V}_P$, obtained by projecting the neuron to vocabulary space. Finally, we prompt GPT-4 to inquire whether a neuron from filtered $N$ promotes a given concept or not, This step is described in Algorithm~\ref{alg: two}, and we include the prompt script in Appendix~\ref{app: neuron-discovery}.

\vspace{1mm}
\noindent \textbf{Neuron Activation }
Following~\citet{geva2022transformer}, we consider a neuron being \emph{activated} in a layer $l$ at a time step $i$ when the neuron's coefficient $m^l_{ij}$ is ranked at top 10. The other alternative would be to define a threshold based on $m^l_{ij}$ to determine its activation. However, coming up with an appropriate threshold poses a challenge, as the threshold value may vary across different layers or even among the individual neurons. Consequently, we opt to focus solely on neurons with the top 10 largest coefficients in our analysis.

\section{Experimental Setup}
% In this section, we describe the experimental setup for interpreting LLM CoT.

\begin{table}[t!]
\centering
\small
\resizebox{0.9\columnwidth}{!}{%
    \begin{tabular}{p{5cm}c}
    \toprule
    \textbf{CoT Prompt} & \textbf{Accuracy}  \\
    \midrule
    CoT & 16.83\%   \\
    w/o Equation (RQ3) & 12.58\% \\
    w/o Textual Explanation (RQ4) & 13.41\%  \\
    AddOnly (RQ5) & 13.26\% \\
    MultOnly (RQ5) & 13.13\% \\
    Incorrect Label (RQ6) & 16.45\% \\
    OOD Label (RQ6) & 7.58\%  \\\bottomrule
    \end{tabular}
}
\caption{The accuracy of Llama2-7B on GSM8k test set based on different CoT prompts.}
\label{tab: accuracy}
\end{table}

\noindent \textbf{Dataset and Model Setup }
We conduct our experiment on the GSM8k dataset \cite{cobbe2021training}, which is widely used for evaluating the arithmetic reasoning capabilities of LLMs. It consists of diverse grade school math word problems and only requires basic arithmetic operations to solve, often involving problem-solving steps ranging from two to eight. We use Llama2-7B \cite{touvron2023llama} as our model to investigate the reasoning capabilities in LLMs. However, we believe that our findings apply to other transformer-based decoder-only LLMs as well. 

% We base our experiments on the CoT prompts obtained from \citet{fu2023chain}, with a slight modification to encourage consistent format in multi-step reasoning, for the ease of further analysis.
Our experiments are based on the CoT prompts obtained from \citet{fu2023chain}, with a slight modification to ensure a consistent format in multi-step reasoning which makes further analysis easier. Each CoT prompt consists of eight exemplars.
% Additionally, we adapt the CoT prompts into different various to reproduce prior observations.
% Additionally, we adapt the CoT prompts according to the methodologies outlined in the respective papers for experiments that involve explaining the prior studies in Section~\ref{sec: result_prior_observation}. 
For reproducibility purposes, we provide a complete list of our prompts in the Appendix \ref{app: prompts}.

Before investigating the mechanism of LLM reasoning, we have conducted experiments to replicate and validate observations made by prior work (Table~\ref{tab:prior_research}). For RQ4 and RQ6, different prior work adopted different ablation designs. We opted for the most suitable and fair design among them. The experimental results based on Llama2-7B are presented in Table~\ref{tab: accuracy}, which present consistent observations as prior research. We refer readers to Appendix~\ref{subsec: replicating-prior-work} for more details.

%%%%%%%%%%%%%%%%%%%%%%%%%%%%%%%%
%% MOVE TO APPENDIX

%%%%%%%%%%%%%%%%%%%%%%%%%%%%%%%%

\noindent \textbf{Summary of Research Questions (RQs) }
Leveraging the lens of neuron activation, we aim to answer two sets of questions. The first set of questions (RQs 1-2) tries to understand the underlying mechanism of LLM reasoning where we initially find different neurons related to arithmetic reasoning and explore the importance of these discovered reasoning neurons for activating reasoning in LLMs. Built upon this foundational understanding of LLMs' reasoning mechanism, the second set of questions (RQs 3-6) attempts to provide a unified explanation of observations made by prior work.

\section{Understanding the Mechanism of Reasoning in LLMs}
\subsection{RQ1: Are there neurons or sub-updates related to the concept of ``reasoning''?}
\label{rq:rq1}

\begin{table*}[t!]
    \centering
    \resizebox{\textwidth}{!}{%
    \begin{tabular}{>{\centering\arraybackslash}p{2cm}>{\centering\arraybackslash}p{4cm}>{\centering\arraybackslash}p{6cm}>{\centering\arraybackslash}p{2.5cm}>
    {\centering\arraybackslash}p{8cm}}
    \toprule
    \textbf{Concept} & \textbf{Seed Tokens} & \textbf{Expanded Tokens}  & \textbf{\#of Neurons} & \textbf{Exemplar Neurons}  \\
    \toprule
    Logical Connectors ($C_{\texttt{logic}}$) & { \{first, so, meaning, therefore, then, next, hence \}} & {\{logic, implies, thus, however, accordingly, subsequently, later, corresponding, etc. \} } & 65  & {L10N9818{\{then, THEN, Then, then, ..\}}, L11N3000{\{therefore, Therefore, accordingly, donc,..\}}, L11N7742, L12N1030} \\
    
    \midrule
    Addition ($C_{\texttt{add}}$) & \{add, addition, +, sum, plus \} & {\{added, U+002B, adding, ++, increment, total, etc.\}  } & 18  & {L12N4814{\{added ,addition ,add,..\}}, L21N7027{\{+ ,add ,U+4e0e, U+306,..\}}, L27N10751{\{+, plus, -, minus,..\}}} \\

    \midrule
    Subtraction ($C_{\texttt{sub}}$) & \{subtract, -, minus, sub \} & \{ -=, negative, U+2212, etc. \} & 2 &  {L19N7900{\{-= ,- ,minus,\u2212, ..\}}, L25N5227} \\

    \midrule
    Multiplication ($C_{\texttt{mul}}$) & \{multiply, product, times, mult, *, x \} & \{ multip, multi, U+00D7, double, twice, triple, fold, larger, etc. \} & 5 &  {L16N10193{\{multip, double, multip, multiply, ..\}}, L18N4462, L20N6554, L22N1345, L22N1236} \\

    \midrule
    Division ($C_{\texttt{div}}$) & \{divide, division, div, /, \% \} & {\{ div, divided, divisions, U+00F7, partition, partitions, etc. \}} & 2 & {L20N10457{\{div ,divided ,division ,U+00F7,.. \}}, L26N1378{\{div, Div, div, Div, division,.. \}}} \\

    \midrule
    Equals to ($C_{\texttt{eq}}$) & \{ =, total, equals, equal, equivalent \}  & {\{ equality, identical, same, exactly, contain, exact, etc. \} } & 6 & {L14N7597{\{identical, difference, differences, equal,..\}}, L18N7531, L18N1850, L20N3177, L20N5535, L24N154} \\

    \midrule
    Calculation ($C_{\texttt{cal}}$) & { \{formula, equation, calculation, algorithm, expression, computation \}} & \{rewrite, sum, application, ratio, percentage, eqn, rate, etc. \} & 14  & {L11N815{\{equation, formula, Formula, diagram,..\}}, L7N7176, L8N3689, L13N2019, L15N3958} \\
     
    \bottomrule
    \end{tabular}
    }
    \caption{List of concepts related to arithmetic reasoning along with their seed tokens and the count of discovered neurons in Llama2-7B. We also list the expanded tokens, promoted by the discovered neurons and their exemplar neurons. For some exemplar neurons, we also show its top-scored vocabulary tokens enclosed within braces. }
    \label{tab:Neurons}
\end{table*}

{To answer this question, we apply the proposed approach in Section~\ref{sec: neuron-discovery-algo} to automatically identify neurons implying a set of 7 concepts, including logical connectors, which plays a crucial role in deciding the reasoning direction, a set of four arithmetic operations (i.e., add, subtract, multiply, and division), and others (equals to and calculation), which are also important to arithmetic reasoning.
% related to arithmetic reasoning. These concepts include (1) Logical Connectors ($C_{logic}$) which are important to decide the reasoning direction after each sentence, (2) Arithmetic concepts ($C_{arith}$) include 4 sub-concepts --- Add ($C_{add}$), Subtract ($C_{sub}$), Multiply ($C_{mul}$), and Division($C_{div}$), and (3) Equals to ($C_{eq}$) and Calculation ($C_{cal}$), which are also concepts important to arithmetic reasoning. 
% Although our list of concepts might not fully encompass arithmetic reasoning, they can help us understand the arithmetic reasoning mechanism in LLMs. 
Though they may not fully encompass arithmetic reasoning, these concepts are sufficient for an initial investigation of neuron activation.
%%%%%%%%%%%%%%%%%%
%% MOVE TO APPENDIX
% In Algorithm~\ref{alg: one}, we randomly select $20$ examples from GSM8k's test set as $\mathcal{E}$ and set $K=20$. Subsequently, we employ Algorithm~\ref{alg: two} to identify associated neurons for each concept, with thresholds $P=20$ and $F=2$. 
%%%%%%%%%%%%%%%%%%
The seed token set $\mathcal{S}_{name}$ for each concept, the identified neuron examples, and the expanded concept tokens found in the identified neurons, are presented in Table~\ref{tab:Neurons}. The specific implementation details are included in Appendix~\ref{app: rq1-details}.
% Specifically, we manually annotate 4-7 seed tokens to filter out irrelevant neurons in $N$ using a threshold $F=2$ and prompt GPT-4 to classify each neuron in the filtered $N$ based on their top $p=20$ dominant tokens.}

% Next, we use Algorithm~\ref{alg: two} to find associated neurons for each concept where we first manually annotate 4-7 seed tokens to filter out irrelevant neurons in $N$ with threshold $F=2$ and finally prompting GPT-4 to classify each neuron in the filtered $N$ based on the top $p=20$ dominant tokens.}

We find a total of 113 neurons associated with the listed concept in Llama2-7B. We performed manual validation of the results and didn't find any objection. {For instance, the ``L21N7027'' neuron, corresponding to the 21st layer and 7027-th row of $V^{21}$, projects with high coefficients to tokens such as ``+'', ``U+4e0e'', ``\&'', ``and'', ``U+acfc'', ``plus'', ``+'', ``AND'', ``U+3068'', etc., and GPT-4 reasonably classified it as a neuron that promotes ``Arithmetic Addition''.} Notably, we discovered neurons that group certain concepts using different language characters. For instance, the ``L21N7027'' neuron promotes tokens like ``and'' and ``+'' with their corresponding translation for Chinese (U+4e0e) and Japanese (U+3068). Additionally, we also found some neurons with somewhat polysemantic characteristics, where a single neuron promotes multiple concepts. For instance, ``L27N10751'' promotes tokens related to both addition (+, plus, +=, ..) and subtraction (-, minus, -+, ..).

% \zyedit{[TODO: @Daking: briefly discuss the results in Table 2. Anything interesting?]} \zy{Add L21N7027 to Table 3. Basically, imagine you are guiding readers to read this table, so it's necessary to make examples using what is included in the table. Same for L27N10751}

\vspace{1mm}
\noindent \textbf{Activation Pattern of Reasoning Neurons}
Our further investigation found out intriguing activation pattern of reasoning neurons throughout an LLM's reasoning process. For example, in Figure~\ref{Fig: connector-dynamics} of Appendix~\ref{app: neuron_dynamics}, we showed that the logical connector neurons are often activated at the beginning of a generated sentence, whereas arithmetic neurons are mostly activated in response to arithmetic symbols and numbers. Once a neuron is activated, it remains activated for a few subsequent time steps. This persistence implies a lasting impact of activated neurons on text generation in its proximity.

\subsection{RQ2: Are the discovered neurons important for eliciting the reasoning capability of LLMs?}
\label{rq:rq2}

% We show the importance of discovered reasoning neurons for LLM to perform reasoning by showing the decrease in their performance when we remove reasoning neurons contribution during inference time. To achieve this, we corrupt the reasoning neurons by adding Gaussian noise to the value vector. 

% To validate the importance of our discovered neurons, we corrupt these neurons by adding Gaussian noise to them. If these neurons are critical to LLM reasoning, the corrupted LLM should present a decrease in their performance.
{To validate the importance of our discovered neurons and assess their \emph{faithfulness} in promoting various reasoning concepts, we perform random noise ablation \cite{meng2022locating} of the discovered neurons. If these neurons are critical to LLM reasoning, the corrupted LLM should exhibit a decrease in performance.}
Specifically, for all reasoning neurons in FF, we added Gaussian noise to the neurons, changing Eqn~\ref{eq:eq3} to $FF^l (x_i^l)= \sum_{j=1}^{d_m} m^l_{ij} (v^l_j + Noise)$. As a baseline, we also corrupted the same number of random neurons for comparison. Subsequently, we run the Llama2-7B with corrupted reasoning neurons and random neurons separately. 
We report the few-shot CoT performance of each LLM variant on the GSM8k test set in Table~\ref{tab: corruption}.
% We report the accuracy, 4.54\% when corrupting reasoning neurons and 11.37\% when corrupting random neurons, as shown in Table~\ref{tab: cuasal_mediation_analysis}.

\begin{table}[t!]
\centering\small
\resizebox{0.8\columnwidth}{!}{%
    \begin{tabular}{lc}
    \toprule
    % \textbf{CoT Setting} & \textbf{Accuracy}  \\
    % \midrule
    % CoT & 16.83\%   \\
    % CoT - reasoning neurons &  4.54\%  \\
    % CoT - random neurons &  11.37\% \\\bottomrule
    \textbf{LLM Variant} & \textbf{Accuracy}  \\
    \midrule
    No corruption & 16.83\%   \\
    w/ corrupted reasoning neurons &  4.54\%  \\
    w/ corrupted random neurons &  11.37\% \\\bottomrule
    \end{tabular}
}
\caption{Llama2-7B's performance before and after corruption of reasoning neurons vs random neurons.}
\label{tab: corruption}
\vspace{-4mm}
\end{table}

We observe a substantial performance decrease of $12.29\%$ when the discovered reasoning neurons are corrupted, in contrast to a decrease of only 5.47\% observed when random neurons are corrupted. The results thus show the essential role of the discovered reasoning neurons in facilitating effective reasoning by LLMs, indicating their \emph{necessity} for eliciting reasoning in LLMs.\footnote{We define a mechanism (e.g., reasoning neuron activation) to be a ``necessity'' for a model capability (e.g., arithmetic reasoning) if the absence of the mechanism results in the model’s inability to demonstrate the capability. 
A relevant concept is ``sufficiency'', which is defined when the model's capability can be attributed to it~\cite{deyoung2019eraser, tuan2021local, rai2023explaining}. We include further discussions in the Limitations section.
% , we define a mechanism to satisfy ``sufficiency'' if the model capability can be solely attributed to it~\cite{deyoung2019eraser}.
} 
% The results thus show the essential role of the discovered reasoning neurons in facilitating effective reasoning by LLMs. 
% Additionally, we discovered that our randomly selected neurons also encompass a small subset of important neurons, as they are activated by CoT prompts, as listed in Table~\ref{tab: neuron_activation}. 
In addition, the performance drop when corrupting random neurons implies that some of these neurons may also play an important role (e.g., for context understanding). As we will show in Table~\ref{tab: neuron_activation}, these neurons reveal non-zero coefficients on average.

\subsubsection{Correlation between the reasoning performance of LLMs and the activation of their reasoning neurons}
\label{subsec: surface_vs_neuron}
Given that the identified neurons make critical contributions to an LLM's arithmetic reasoning, a natural question is: Does an LLM's reasoning performance correlate positively with how their reasoning neurons are activated? To answer this question, we performed an experiment in the zero-shot CoT setting~\citep{kojima2022large}.
% We further explore the relationship between the reasoning ability of Llama2-7B and the activation of discovered reasoning neurons under a zero-shot CoT setting. 
We specifically selected a zero-shot CoT setting for this analysis because it is unbiased due to the lack of demonstration. In our experiment, we select four zero-shot CoT prompts with varying levels of accuracy on the GSM8K test set, sourced from \citet{yang2023large}. The prompts include \textit{``Let's think step by step''}, \textit{``Take a deep breath and work on this problem step-by-step''}, \textit{``Break this down''}, and \textit{``A little bit of arithmetic and a logical approach will help us quickly arrive at the solution to this problem''}. Their respective accuracies are 7.05\%, 4.47\%, 11.06\%, and 5.83\% in Llama2-7B. In Figure~\ref{Figure 1: surface_vs_neuron}, we plot their accuracy along with the average coefficient of their reasoning neurons per time step during the output generation. The result confirms our hypothesized positive correlation. It also reveals the potential of \emph{predicting} an LLM's reasoning performance by examining the activation of their reasoning neurons, without needing human-annotated labels. We leave systematic explorations of this potential to the future.

\begin{figure}[t!]
\centering
\includegraphics[width=\columnwidth] {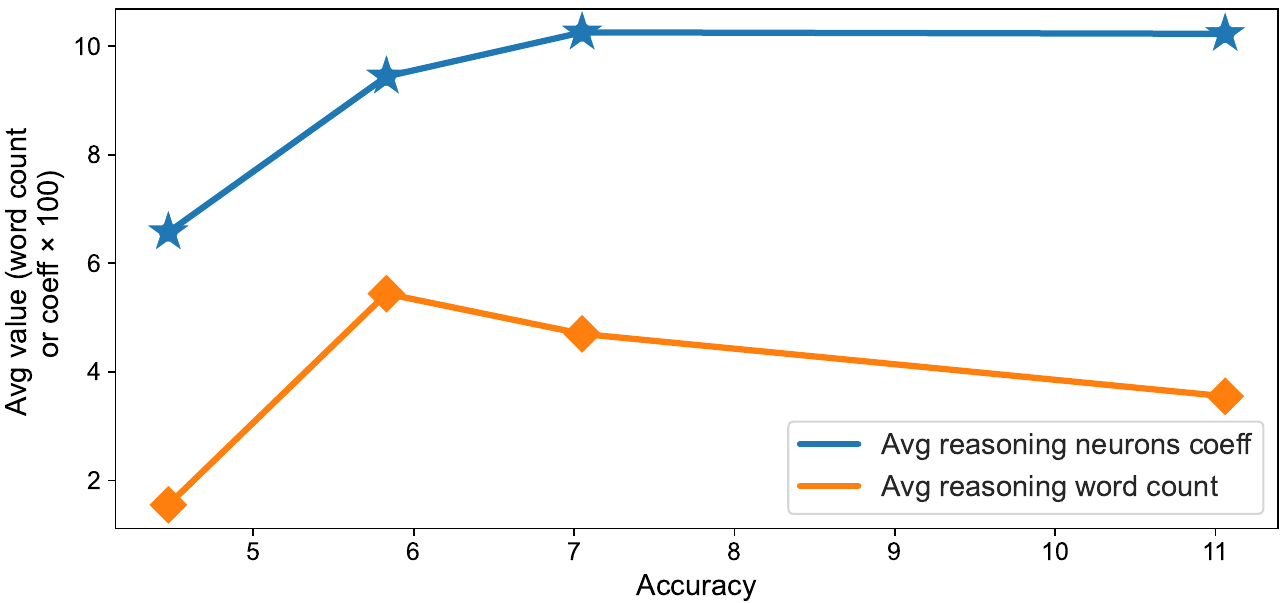} 
\caption{Correlation between prompt accuracy and the LLM's average coefficient on the discovered reasoning neurons (blue stars) and its average count of reasoning tokens (orange diamonds).
% The correlation between prompt accuracy on the GSM8K test set and the count of reasoning words versus prompt accuracy and the average coefficient value of reasoning neurons per generation time step. To facilitate visualization, the average coefficient values are divided by 100 to fit within the plot.
}
\label{Figure 1: surface_vs_neuron}
% \vspace{-4mm}
\end{figure}

Additionally, we examine if the same correlation can be observed superficially at the word level, because, if the word-level statistics present the same correlation, it could be a more convenient approach of probing into an LLM's reasoning performance than neuron activation. To respond to this question, we similarly examine the correlation between the count of reasoning tokens in the LLM generation, using a combination of the human-annotated seed tokens and the GPT-4-extracted expanded tokens listed in Table~\ref{tab:Neurons}, and their accuracy on the GSM8k test set. Our result is presented in Figure~\ref{Figure 1: surface_vs_neuron}. Intriguingly, we observe no positive correlation between the two factors, which thus highlights the importance of performing neuron-level analysis, as the latter offers direct insights into the functioning of LLMs that may not be visible from simply analyzing their superficial text generation.

\section{Understanding Prior Observations via the Lens of Neuron Activation} \label{sec: result_prior_observation}

In this section, we revisit the major findings from prior work and use the activation of FF neurons in an LLM to explain them. For each research question (RQ), our analysis will be based on how each CoT prompt variant triggers different neuron activation patterns. These observations are summarized in Table~\ref{tab: neuron_activation}, where the number of total or unique activated neurons is counted across the encoding steps of each CoT prompt, and the reported coefficient is an average per neuron. To show a baseline, we also report the average coefficient of randomly sampled neurons.

\begin{table*}[t!]
    \centering
    \resizebox{\textwidth}{!}{%
    \begin{tabular}{>{\centering\arraybackslash}p{3cm}|>{\centering\arraybackslash}p{2.5cm}|>{\centering\arraybackslash}p{2.5cm}>{\centering\arraybackslash}p{2.5cm}>{\centering\arraybackslash}p{2.5cm}>{\centering\arraybackslash}p{2.5cm}|>{\centering\arraybackslash}p{2.5cm}>{\centering\arraybackslash}p{2.5cm}|>{\centering\arraybackslash}p{2.5cm}|>{\centering\arraybackslash}p{1.5cm}}
    \toprule
    \textbf{Prompt Type} & \textbf{Logical Connectors ($C_{logic}$)} & \textbf{Addition ($C_{add}$)}  & \textbf{Subtraction ($C_{sub}$)} & \textbf{Multiplication ($C_{mul}$) } & \textbf{Division \newline ($C_{div}$)} & \textbf{Equals to \newline ($C_{eq}$)} & \textbf{Calculation ($C_{cal}$)} & \textbf{Reasoning Neurons \newline} (Total) & \textbf{Random Neurons}  \\
    \toprule
    CoT & (226, 35, 2.62) & (599, 15, 2.54) & (87, 2, 2.38) & (98, 4, 3.075) & (28, 2, 2.19) & (19, 4, 2.27) & (62, 6, 1.37) & (1119, 68, 2.51) & 1.15  \\
    
    \midrule
    w/o Equation (RQ3)  & (207, 36, 2.60) & (449, 13, 2.18) & (41, 2, 2.09) & (67, 4, 2.94) & (19, 2, 1.71) & (11, 3, 2.14) & (48, 6, 1.45) & (842, 66, 2.29) & 1.53  \\

    \midrule
    w/o Textual Explanation (RQ4) & (85, 28, 1.50) & (450, 13, 2.94) & (86, 2, 2.34) & (73, 4, 2.95) & (51, 2, 2.46) & (11, 3, 2.68) & (27, 6, 1.03) & (783, 58, 2.62) & 1.45  \\

    \midrule
    AddOnly (RQ5)  & (286, 32, 2.55) & (651, 14, 3.04) & (97, 2, 2.62) & (173, 4, 2.24) & (13, 1, 1.66) & (40, 5, 2.24) & (90, 8, 1.25) & (1350, 66, 2.65) & 1.39    \\

    \midrule
    MultOnly (RQ5)  & (212, 25, 2.37) & (322, 11, 2.37) & (97, 2, 1.82) & (229, 5, 2.7) & (36, 1, 1.87) & (28, 4, 2.79) & (127, 10, 1.37) & (1051, 58, 2.17) & 1.5 \\

    \midrule
    Incorrect Label (RQ6) & (221, 35, 2.60) & (601, 15, 2.52) & (97, 2, 2.41) & (103, 4, 2.86) & (28, 2, 2.10) & (24, 4, 2.26) & (65, 6, 1.35) & (1139, 68, 2.47) & 1.45  \\

    \midrule
    OOD Label (RQ6) & (240, 41, 2.53) & (543, 15, 2.62) & (92, 2, 2.20) & (97, 4, 2.94) & (26, 2, 2.21) & (23, 4, 2.02) & (66, 7, 1.39) & (1087, 75, 2.50) & 1.50 \\

    \bottomrule
    \end{tabular}
    }
    \caption{For each prompt variant, we present (count of activated neurons, count of unique activated neurons, average coefficient) for each concept or total. We also present the average coefficient of random neurons as a baseline.
    % Number of activated reasoning and random neurons and their average coefficient (count of activated neurons, count of unique activated neurons, average coefficient) for different CoT prompts, as described in Table~\ref{tab:prior_research}.
    }
    \label{tab: neuron_activation}
    % \vspace{-2mm}
\end{table*}

\subsection{RQ3: Why do equations matter?}
\label{RQ3}

Prior works \cite{wang-etal-2023-towards, ye-etal-2023-complementary, madaan2022text} have shown that equations play an important role in eliciting reasoning in LLMs. 
% They demonstrate this by showing a decrease in LLM performance when equations are omitted from the CoT, as shown in Table~\ref{tab: accuracy}. We investigate the underlying reasons for this observation by examining the disparities in the activation of identified reasoning neurons in Llama2-7B with and without equation CoT prompts, as shown in Table~\ref{tab:prior_research}. We list the number of activated reasoning neurons and their average coefficient for both CoT prompts in Table~\ref{tab: neuron_activation}. 
Looking into its activation pattern, we found the CoT prompt without equations (denoted as ``w/o Equation'') activates fewer reasoning neurons overall, 842 activations, compared to the CoT prompt with equations (CoT), 1119 activations. Furthermore, 
% upon examining the activation of neurons associated with individual concepts within reasoning neurons, as outlined in Table~\ref{tab:Neurons}, 
we observed a decrease in both the number of activated neurons for individual concepts and their corresponding average coefficients across all categories.
% when comparing the "w/o equation" prompt to the "CoT" prompt. 
This shows that equations play an important role in activating the reasoning neurons which are deemed to be important for arithmetic reasoning. As a result, the presence of equations can help elicit arithmetic reasoning in LLMs.

Interestingly, we also note that although there were no equations or arithmetic operators in the ``w/o Equation'' prompt, neurons associated with arithmetic operations (i.e., $C_{add}$, $C_{sub}$, $C_{mul}$, $C_{div}$) were still activated. This indicates that even in the absence of explicit equations in the CoT prompt, LLMs are capable of recognizing the necessity of performing arithmetic operations, which explains the 12.58\% retained accuracy in Table~\ref{tab: accuracy}.

\subsection{RQ4: Why do text explanations matter?}
\label{RQ4}

% Prior works \cite{wang-etal-2023-towards, ye-etal-2023-complementary, madaan2022text} show that textual explanations are another crucial part of the CoT prompt for eliciting reasoning in LLMs. 
% The absence of textual explanations in CoT prompts leads to a decrease in performance, as illustrated in Table~\ref{tab: accuracy}. To investigate this observation, we analyze the difference in the activation of identified reasoning neurons in Llama2-7B for prompt with and without textual explanation, as shown in Table~\ref{tab:prior_research}. We list the number of activated reasoning and their average coefficient for both CoT prompts in Table~\ref{tab: neuron_activation}. 
% This 
The importance of textual explanations as found in prior work \cite{wang-etal-2023-towards, ye-etal-2023-complementary, madaan2022text} is also consistent with our observation. We found that the CoT prompt without textual explanations activates reasoning neurons fewer times, 783 activated neurons, compared to the CoT prompt with explanation, 1119 activated neurons. Specifically, we observe a significant decrease in the activation of neurons associated with logical connectors ($C_{logic}$) and a slight decrease in the activation of neurons associated with arithmetic operations (particularly $C_{add}$). This shows the utility of textual explanations not only in activating neurons associated with logical connectors, crucial for determining the reasoning direction but also in activating neurons associated with arithmetic operations.

% \subsection{RQ4: Why arithmetic diversity in exemplars doesn't matter?}
\subsection{RQ5: Why does arithmetic diversity in exemplars matter?}
\label{RQ5}

\citet{ye-etal-2023-complementary} showed that arithmetic diversity in exemplars is important for arithmetic reasoning, i.e. CoT prompts that consist of all arithmetic operations in their demonstrations yield better performance than the ones that do not. 
% Specifically, \citet{ye-etal-2023-complementary} showed that CoT prompts that consist of only addition or multiplication have lower performance than the CoT prompt that consists of both arithmetic operations. We speculate that the performance decline is because when we only have one type of arithmetic operation in the CoT prompt, we bias LLM towards favoring that specific arithmetic operation in problem-solving, which can often lead to incorrect reasoning paths. To explore this hypothesis, we analyze the activation of identified reasoning neurons in Llama2-7B across three different prompt types, AddOnly (solely addition), MultOnly (solely multiplication), and CoT (containing both operations), as shown in Table~\ref{tab:prior_research}. We list the number of activated reasoning neurons and their average coefficient for all CoT prompts in Table~\ref{tab: neuron_activation}.
Our results in Table~\ref{tab: neuron_activation} indicate that the performance decline is likely caused by the bias introduced by the partial operators.
% because when we only have one type of arithmetic operation in the CoT prompt, we bias LLM towards favoring that specific arithmetic operation in problem-solving, which can often lead to incorrect reasoning paths.
% To verify this hypothesis, we assess the activation levels of both AddOnly and MultOnly prompts in comparison to the CoT prompt, which we regard as a better activation in terms of both the number of activations and their average coefficients. 
We observe that the AddOnly prompt activates a higher number of $C_{add}$ neurons (651 vs 599) and $C_{mul}$ neurons (173 vs 98) when compared to CoT, but fewer $C_{div}$ neurons with a lower average coefficient. Similarly, we found that MultOnly activates a significantly higher number of $C_{mul}$ neurons when compared to the CoT prompt (229 vs 98), but significantly fewer $C_{add}$ neurons (322 vs 599). This shows that although both AddOnly and MultOnly activate the neurons related to arithmetic reasoning, they exhibit a bias toward emphasizing specific arithmetic operations, which explains their degraded performance.

\subsection{RQ6: Why does incorrect reasoning or gold label not matter?}
\label{RQ6}

Prior work \cite{wang-etal-2023-towards, min2022rethinking} shows that incorrect labels in the few-shot exemplars do not matter, as long as the labels come from the same distribution. Consistent with our previous findings, we observed a similar reasoning neuron activation pattern for CoT prompts with correct and incorrect labels. However, despite a $9.25\%$ decrease in accuracy for the ``OOD Label'' prompt, it still exhibited a similar reasoning neuron activation pattern compared to the patterns of CoT. 
% This discrepancy contradicts our prior findings that indicate a positive correlation between reasoning neuron activation and LLMs' arithmetic reasoning.

To understand this phenomenon, we conducted the second analysis. In the prior work, \citet{geva2022transformer} found that LLMs refresh their token representations by accumulating sub-updates (Section~\ref{sec:sec2.2}). Therefore, two CoT prompts with similar performance presumably should reveal similar sub-updates per layer in the corresponding step, and vice versa. To validate it, we looked into the neuron activation for each prompt in the encoding steps where the labels were manipulated (e.g., the positions of ``1'' and ``Dawson'' in Table~\ref{tab:prior_research}), as other input tokens are the same in all the three prompts. We then plotted the overlap of activated neurons per layer between CoT and ``Incorrect Labels'' or between CoT and ``OOD Label'' in Figure~\ref{app-fig: incorrect_correct_label} of Appendix~\ref{app: neruon-overlap}. Note that here we consider all activated neurons, no matter if they are discovered as reasoning neurons or not. We observe a substantial overlap of $63.05\%$ on average in the former case while merely $14.91\%$ in the latter. The observation is consistent with our hypothesis, showing that the activation of FF neurons can be used to explain the performance of CoT prompting.

The two observations (i.e., inconsistent reasoning neuron's activation pattern based on Table~\ref{tab: neuron_activation} but consistent sub-update pattern based on the overlap analysis) thus imply that the activation of reasoning neurons are \emph{necessary} but not \emph{sufficient} to elicit reasoning in LLMs. In fact, our qualitative analysis showed that in the case of providing OOD labels, the LLM still engages in reasoning, and their reasoning paths are similar to those prompted by correct labels (see examples in Appendix~\ref{app: example-ood}), which explains the activation of their reasoning neurons. However, this reasoning is biased by the use of OOD tokens as variables, leading to messy variable references and an increasing amount of incorrect reasoning as the reasoning proceeds. We include a further discussion in Limitations.

\section{Conclusions}
Our work is among the first in applying neuron activation analysis to understanding LLMs in arithmetic reasoning. Our results offer valuable insights into the role of neurons and their utility in understanding the internal mechanism of LLMs. We thus expect this work to pave the way for future research on LLM interpretability.

\section*{Limitations}

\paragraph{Sufficiency vs Necessity}
% A crucial question to raise is, \emph{does neuron activation represent all about LLM reasoning?} {In other words, is an analysis of neuron activation \emph{sufficient} to completely explain the LLM reasoning capability?}. Despite its efficacy in explaining RQs 3-6, it is inherently limited by its focus on analyzing neurons individually without considering the interaction among neurons or other LLM components (e.g., attention modules). Specifically, a major limitation of our approach is that it may not be suitable for understanding very complex phenomena that arise from the interactions among different components of LLMs. For instance, to fully understand in-context learning within CoT, analyzing neurons in isolation may prove insufficient. Instead, as explored by \citet{olsson2022context}, it requires studying attention heads and their circuits, which are sub-networks of neurons. \daking{This shows that the activation of reasoning neurons is \emph{necessary} but not \emph{sufficient} to elicit the reasoning ability of LLMs.} Despite this limitation, through our study, we show that analysis of neuron activation can play an important role. Therefore, future work should study it together with other approaches such as circuit analysis \cite{olsson2022context, wang2022interpretability}, top-down-approach \cite{zou2023representation, meng2022locating}, etc. to provide a more complete picture of LLMs' inner mechanism for reasoning.

We show that neuron activation is a \emph{necessary} condition for LLM to elicit reasoning capability through random noise ablation study in RQ2. However, another crucial question to raise is, \emph{does neuron activation represent all about LLM reasoning?} In other words, is an analysis of neuron activation \emph{sufficient} to completely explain the LLM reasoning capability? As discussed in Section~\ref{RQ6}, although CoT with incorrect OOD labels has lower accuracy than CoT with correct labels (16.83\% vs 7.58\%), they show a similar number of reasoning neuron activations (1119 vs 1087) and similar average coefficients (2.51 vs 2.50). This indicates that the activation of reasoning neurons is \emph{necessary} but not \emph{sufficient} to elicit the reasoning ability of LLMs.  
% Furthermore, our analysis doesn't consider the interaction among neurons and other LLM components (e.g., attention modules) and behaviors that arise from the interactions among different components of LLMs. 
Despite its efficacy in explaining RQs in this work, the analysis of neuron activation is inherently limited by its focus on analyzing neurons individually, without considering the interaction among neurons or other LLM components (e.g., attention modules); as a result, it may not be able to explain complicated model behaviors that result from the interactions among different components in an LLM.
For instance, to understand in-context learning within CoT, analyzing neurons in isolation may prove insufficient. Instead, as explored by \citet{olsson2022context}, it requires studying attention heads and their circuits, which are sub-networks of neurons. Despite this limitation, through our study, we show that analysis of neuron activation can play an important role. Therefore, future work should study it together with other approaches such as circuit analysis \cite{olsson2022context, wang2022interpretability}, top-down approach \cite{zou2023representation, meng2022locating}, etc., to provide a more complete picture of LLMs' inner mechanism for reasoning.

\paragraph{Limitations of pre-defined concepts}
Although we employ seven concepts introduced in Section~\ref{rq:rq1} to study arithmetic reasoning in LLMs, they may not represent the full scope of arithmetic reasoning. Hence, our study is also limited to the scope of these seven concepts. Furthermore, the activation of these neurons may only indicate the appearance of these concepts during an LLM's reasoning process, but this can be easily ``faked'' (e.g., prompting an LLM to produce a sequence of concept tokens pretending to be performing reasoning). As a result, the coefficient of reasoning neurons as a metric is more helpful when the prompts to LLMs are valid. Thus, it is important to exercise caution when drawing conclusions from the analysis.

\paragraph{Generalization}
Our analysis in this paper was performed on only Llama2-7B. Therefore, there is a concern about whether the insights we observed generalize to other LLMs. To answer this question, we conducted a preliminary study based on Llama3-8B~\cite{llama2024}. In Table~\ref{tab: llama3-neurons}, we present the discovered reasoning neurons when employing our proposed neuron discovery algorithm in Section~\ref{sec: neuron-discovery-algo}. From the results, we confirm that reasoning neurons do exist in various LLMs. However, during the RQ2 investigation, we have found that Llama3-8B behaved very differently to Llama2-7B. Specifically, it is highly sensitive to random noise ablation \cite{meng2022locating}. Even adding a small noise to ablate a few random neurons (approximately 10-20) can drastically decrease its performance from 45.23\% to less than 2.00\%. We believe this sensitivity is due to the model activations being thrown off-distribution by the addition of the noise \cite{chan2022causal}. 
This observation suggests that different approaches may be needed to evaluate the mechanism of different LLMs. More systematic studies should be conducted to investigate these limitations in the future.

\section*{Ethics Statement}
We do not anticipate any severe ethical issues from using the proposed approach. We use fully open-sourced datasets and will open-source our results and dataset as well. On the other hand, we stress the positive impact of our work, as it contributes to interpreting the black box of LLMs. Forming a clear understanding of the inner mechanism of LLMs is crucial for their safe and trustworthy applications. With our investigation of neuron activation for understanding LLMs, we hope to inspire more researchers to extend the research of LLM interpretability. It is also our plan to connect neuron activation with the present societal concerns around LLM safety (e.g., analyzing an LLM's reasoning process and detecting potential vulnerabilities through their neuron activation patterns).

\section*{Acknowledgements}
This project was sponsored by the National Science Foundation (SHF2311468) and College of Computing and Engineering and the Department of Computer Science at George Mason University. This project was also supported by resources provided by the Office of Research Computing at George Mason University (\url{https://orc.gmu.edu}) and funded in part by grants from the National Science Foundation (2018631). We thank the anonymous reviewers and members at GMU NLP group for their feedback on this work.

\bibliography{acl2024}
\bibliographystyle{acl_natbib}

\appendix

\section{Prompt for Neuron Annotation with GPT-4} \label{app: neuron-discovery}

To implement the $\texttt{GPT4ConceptQuery}$ function in Algorithm~\ref{alg: two}, we query GPT-4 using the following prompt: \textit{``A neuron in language model promotes the following set of words: $w_1, .., w_{P}$. Is this neuron promoting $C_{name}$? First, answer in Yes or No format and provide an explanation.''} The function returns ``Yes'' when GPT-4 considers the neuron (as represented by their projected vocabulary tokens) to represent the target concept $C_{name}$. We additionally prompt GPT-4 to provide an explanation as it empirically motivates more precise results from GPT-4.

\section{Additional Details of Replicating Observations of Prior Work}
\label{subsec: replicating-prior-work}
Before investigating the mechanism of LLM reasoning, we first conduct experiments to replicate and validate observations made by prior work (Table~\ref{tab:prior_research}). The experimental results based on Llama2-7B are presented in Table~\ref{tab: accuracy}. We successfully replicated all the results of the prior work.

Although some research questions (RQs) were common in prior work, the experiment design could differ. In these cases, we opted for a more suitable or fair experiment design among them. Specifically, for RQ4, ``Does textual explanation matter?'', we follow the specification of \citet{ye-etal-2023-complementary} instead of \citet{madaan2022text}. \citet{madaan2022text} ablated the text and rewrites the multiple equations into a single equation to evaluate the importance of the text. We find it unfair to compare the importance of equations in the few-shot exemplar as single problem-solving steps rather than multiple steps. In our experiments, we only remove text while retaining all the equations from our original CoT instead of restructuring them into singular equations. Similarly, for RQ6, ``Does correct reasoning or gold label matter?", \citet{ye-etal-2023-complementary} proposed to manipulate only the labels of the equation. On the other hand, \citet{wang-etal-2023-towards} proposed to manipulate other components such as operators and textual explanations as well. We follow the specification of \citet{ye-etal-2023-complementary} instead of \citet{wang-etal-2023-towards} for its simplicity and ease of analysis.

\section{Additional Implementation Details for Neuron Discovery (RQ1) } \label{app: rq1-details}
In Algorithm~\ref{alg: one}, we randomly select $20$ examples from the GSM8k \cite{cobbe2021training} test set as $\mathcal{E}$ and set $K=20$. Additionally, we perform simple greedy decoding on Llama2-7B that consists of 7 billion parameters using a single NVIDIA A100 GPU for 6-7 hours to save the candidate neurons using Algorithm~\ref{alg: one}. Subsequently, we employ Algorithm~\ref{alg: two} to identify associated neurons for each concept, with thresholds $P=20$ and $F=2$ where we prompt GPT-4 $\sim 1300$ times to obtain the reasoning neurons listed in Table~\ref{tab:Neurons}.

\section{Reasoning Neurons Activation Dynamics} \label{app: neuron_dynamics}
To better understand the activation pattern of identified reasoning neurons in Section~\ref{rq:rq1}, we plot their activation throughout an LLM's reasoning text for a randomly selected example, as shown in Figure~\ref{Fig: arithmetic-dynamics} and Figure~\ref{Fig: connector-dynamics}. Our goal is to discern the activation sites of these reasoning neurons and utilize this information to understand the role of these reasoning neurons in each reasoning step or process. To this end, we first divide the LLM's reasoning text into four sections to simplify the observation - (1) Beginning of a sentence (BOS) (2) Equations (3) Numbers (4) Other texts. The activation showed a clear pattern of activation for both neurons related to arithmetic operations and logical connections. In Figure~\ref{Fig: arithmetic-dynamics}, the heightened activation of arithmetic neurons, encompassing those involved in addition, subtraction, multiplication, and division, within equations is evident. Conversely, Figure~\ref{Fig: connector-dynamics} demonstrates increased activation of logical connection neurons at the beginning of sentences (BOS). These observations underscore the specific roles played by different neurons in the reasoning process.

\begin{figure*}[h]
\includegraphics[width= \textwidth] {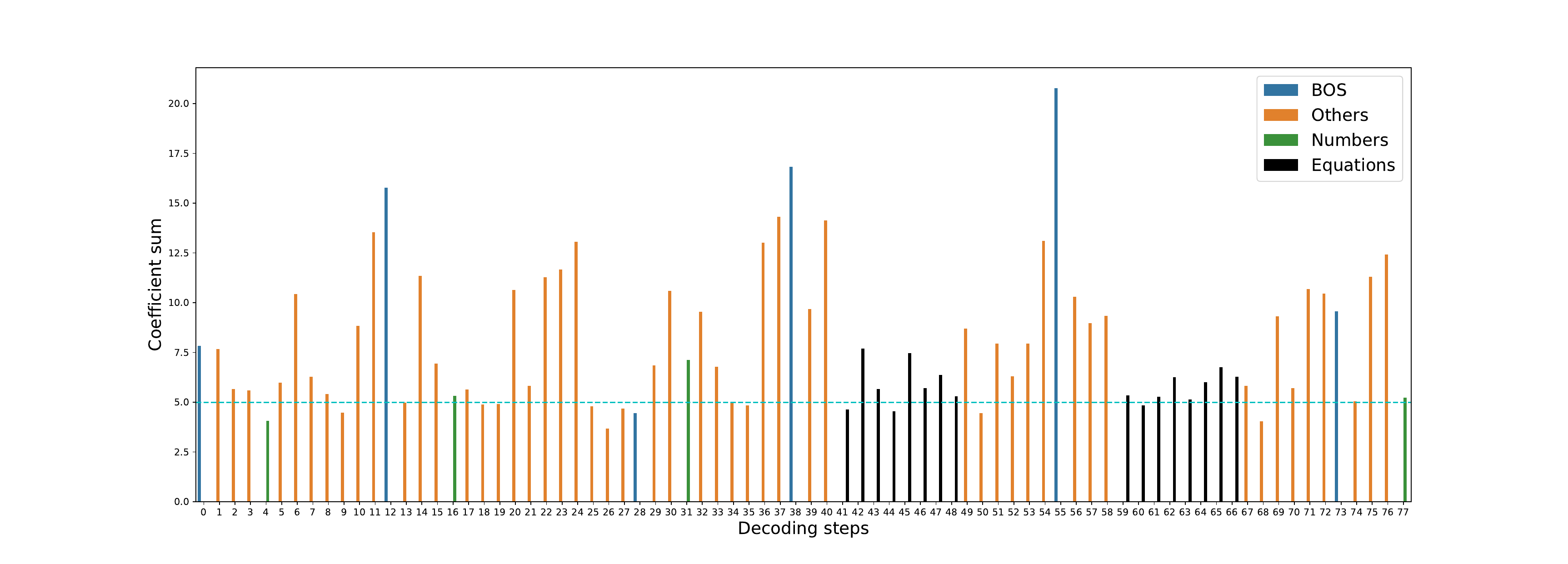} 
\caption{Activation pattern of logical connector neurons for a randomly sampled example. The horizontal dotted line represents the average coefficient of randomly sampled neurons for the same set of examples.}
\label{Fig: connector-dynamics}
\centering
\end{figure*}

\begin{figure*}[h]
\includegraphics[width= \textwidth] {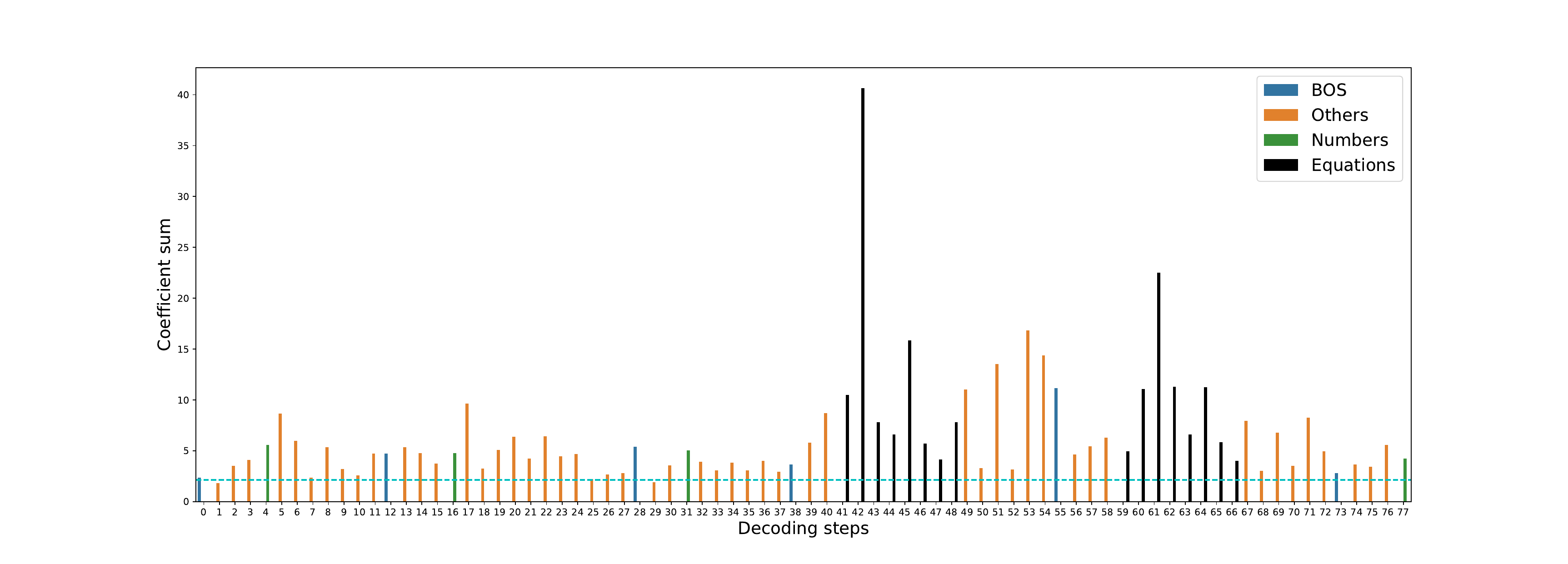} 
\caption{Activation pattern of neurons related to arithmetic neurons (add, subtraction, multiplication, division) for a randomly sampled example. The horizontal dotted line represents the average coefficient of randomly sampled neurons for the same set of examples.}
\label{Fig: arithmetic-dynamics}
\centering
\end{figure*}

\section{Neuron Activation Overlap between CoT with Correct Labels and Incorrect or OOD Labels}  \label{app: neruon-overlap}
The neuron activation overlap between CoT prompt with correct labels and incorrect or OOD labels, discussed in Section~\ref{RQ6} is shown in Figure~\ref{app-fig: incorrect_correct_label}.
\begin{figure}[h!]
\centering
\includegraphics[width=0.9\linewidth] {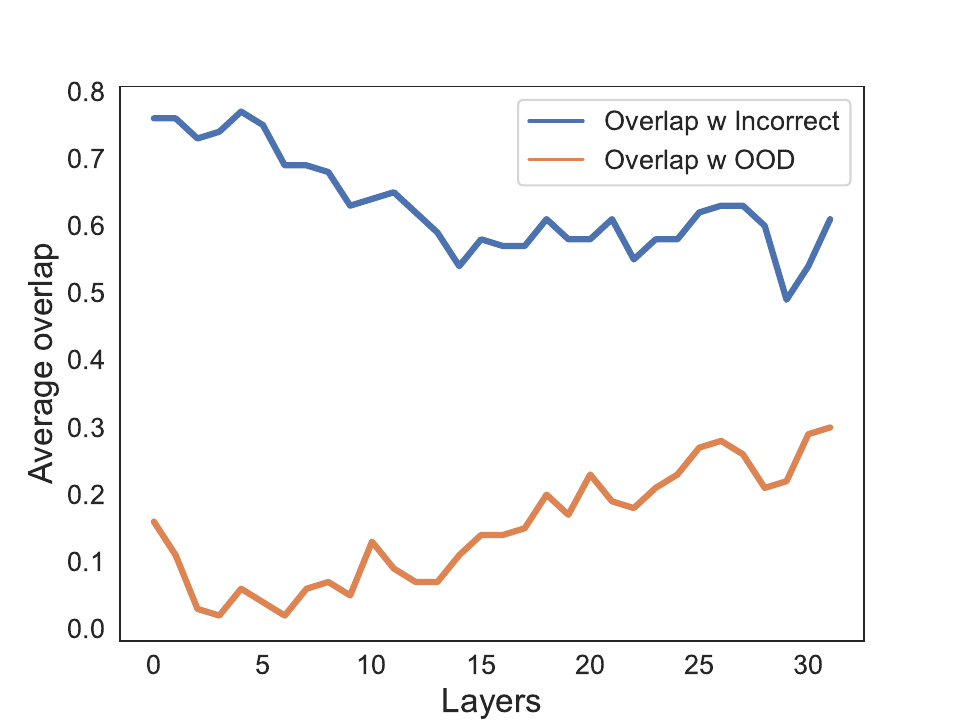}
\caption{Overlap of neuron activation when the LLM is prompted with CoT and ``Incorrect Labels'' or ``OOD Labels''.}
\label{app-fig: incorrect_correct_label}
\centering
\end{figure}

\section{Example Predictions for CoT with OOD labels} \label{app: example-ood}
We list the example prediction for CoT with OOD labels as discussed in Section~\ref{RQ6} is listed in Table~\ref{app-tab: ood_cot}.

% % \input{appendices/ood_labels_inference}
% \begin{table*}[]
% \centering
%     \begin{tabular}
    
%     Question: Janet's ducks lay 16 eggs per day. She eats three for breakfast every morning and bakes muffins for her friends every day with four. She sells the remainder at the farmers' market daily for \$2 per fresh duck egg. How much in dollars does she make every day at the farmers' market?
%     \end{tabular}
% \caption{ Example prediction for CoT prompts with out-of-distribution (OOD) labels}
% \label{app-tab: ood-inference}
% \end{table*}

\begin{table*}
  \centering
  \resizebox{\textwidth}{!}{%
  \begin{tabular}{l}
    \hline
    Question: Janet's ducks lay 16 eggs per day. She eats three for breakfast every morning and bakes muffins for her friends \\ every day with four. She sells the remainder at the farmers' market daily for \$2 per fresh duck egg. How much in dollars does \\ she make every day at the farmers' market? \\
    Let's think step by step \\
    First Janet's ducks lay 16 eggs per day.\\ 
    Then she eats 3 for breakfast every day.\\
    So she has 16 - 3 = Dylan eggs left.\\
    Then she bakes muffins for her friends every day with 4.\\
    So she has Dylan - 4 = Dylan eggs left.\\
    Then she sells the remainder at the farmers' market daily for \$2 per fresh duck egg.\\
    So she has Dylan - 4 = Dylan eggs left.\\
    So she makes \$2 * Dylan = Dylan dollars.\\
    The answer is Dylan. \\
    \\

    Question: Two trains leave San Rafael at the same time. They begin traveling westward, both traveling for 80 miles. The next \\ day, they travel northwards, covering 150 miles. What's the distance covered by each train in the two days?\\
    Let's think step by step\\
    First two trains leave San Rafael at the same time.\\
    Then they begin traveling westward, both traveling for 80 miles.\\
    So they travel 80 miles each.\\
    Then the next day, they travel northwards, covering 150 miles.\\
    So they travel 150 miles each.\\
    So they travel 80 + 150 = Dylan miles each.\\
    The answer is Dylan. \\
    \hline
  \end{tabular}
  }
  \caption{\label{citation-guide}
     Example prediction for CoT prompts with out-of-distribution (OOD) labels
  }
\end{table*}

\section{Implementation Details}
We use models, Llama2-7B \cite{touvron2023llama} and GPT-4 \cite{openai2023gpt}, and the GSM8K dataset \cite{cobbe2021training} that are consistent with their intended use. For each experiment in our analysis (RQs 2-6), we perform simple greedy decoding on Llama2-7B which consists of 7 billion parameters using a single NVIDIA A100 GPU for 6-7 hours.

\section{Neuron Discovery on Llama3-8B}
We apply our neuron discovery approach, introduced in Section~\ref{sec: neuron-discovery-algo}, to automatically discover neurons for a pre-defined set of seven concepts. We find a total of 112 neurons associated with the listed concept in Llama3-8B as listed in Table~\ref{tab: llama3-neurons}.

\begin{table*}[t!]
    \centering
    \resizebox{\textwidth}{!}{%
    \begin{tabular}{>{\centering\arraybackslash}p{2cm}>{\centering\arraybackslash}p{4cm}>{\centering\arraybackslash}p{6cm}>{\centering\arraybackslash}p{2.5cm}>
    {\centering\arraybackslash}p{8cm}}
    \toprule
    \textbf{Concept} & \textbf{Seed Tokens} & \textbf{Expanded Tokens}  & \textbf{\#of Neurons (Llama2-8B)} & \textbf{Exemplar Neurons (Llama2-8B)}  \\
    \toprule
    Logical Connectors ($C_{\texttt{logic}}$) & { \{first, so, meaning, therefore, then, next, hence \}} & {\{logic, implies, thus, however, accordingly, subsequently, later, corresponding, etc. \} } & 54 & {L11N10716{therefore, Therefore, thus}, L11N3000{\{then, entonces, Then,..\}}} \\
    
    \midrule
    Addition ($C_{\texttt{add}}$) & \{add, addition, +, sum, plus \} & {\{added, U+002B, adding, ++, increment, total, etc.\}  }  & 20 & {L15N2531{\{sum ,-total ,total, ..\}}, L17N11677{\{aggregate, cumulative, u+5408, Sum, ..\}}} \\ 

    \midrule
    Subtraction ($C_{\texttt{sub}}$) & \{subtract, -, minus, sub \} & \{ -=, negative, U+2212, etc. \}  & 6 &  {L18N10635{\{minus, \_minus, plus, ..\}}, L19N13312{\{subtract, Subtract, subtraction, ..\}}} \\

    \midrule
    Multiplication ($C_{\texttt{mul}}$) & \{multiply, product, times, mult, *, x \} & \{ multip, multi, U+00D7, double, twice, triple, fold, larger, etc. \}  & 6 & {L17N1738{\{\-times ,multiplied ,times, ..\}}, L18N11474, L18N5099, L21N8702, L22N9035} \\

    \midrule
    Division ($C_{\texttt{div}}$) & \{divide, division, div, /, \% \} & {\{ div, divided, divisions, U+00F7, partition, partitions, etc. \}} &  8 & {L14N8572{\{dividing, este, imo, .. \}}, L17N1507{\{recip, reciprocal, Division, .. \}}} \\

    \midrule
    Equals to ($C_{\texttt{eq}}$) & \{ =, total, equals, equal, equivalent \}  & {\{ equality, identical, same, exactly, contain, exact, etc. \} }  & 8 & {L22N12005{\{=, =, =true, +=, ..\}}, L25N8438, L25N14241, L26N2189, L26N9086, L27N4442} \\

    \midrule
    Calculation ($C_{\texttt{cal}}$) & { \{formula, equation, calculation, algorithm, expression, computation \}} & \{rewrite, sum, application, ratio, percentage, eqn, rate, etc. \}  & 10  & {L16N4877{\{formula, formulas, Formula, CALC, ..\}}, L15N9238, L17N10902, L21N237, L21N10940} \\
     
    \bottomrule
    \end{tabular}
    }
    \caption{List of concepts related to arithmetic reasoning along with their seed tokens and the count of discovered neurons in Llama3-8B. We also list the expanded tokens, promoted by the discovered neurons and their exemplar neurons. For some exemplar neurons, we also show its top-scored vocabulary tokens enclosed within braces. }
    \label{tab: llama3-neurons}
\end{table*}

\section{CoT Prompts for Reproducibility}\label{app: prompts}
We list all the CoT prompts used in our analysis, RQs 1-6. The CoT prompts (Correct) is listed in Table \ref{app-tab: cot-prompt}, CoT prompt w/o equation is listed in  Table \ref{app-tab: without_equation}, CoT prompt w/o text is listed in Table \ref{app-tab: without-text}, AddOnly prompt is listed in Table~\ref{app-tab: add-only}, MultOnly prompt is listed in Table~\ref{app-tab: mult-only}, CoT prompt with incorrect labels is listed in Table \ref{app-tab: incorrect-cot}, and CoT prompt with OOD labels is listed in Table \ref{app-tab: ood_cot}.

% Appendices prompt ---------------------------------------------------------
\label{app: prompts}
\begin{table*}
  \centering
  \resizebox{\textwidth}{!}{%
  \begin{tabular}{l}
    \hline
    Question: There are 15 trees in the grove. Grove workers will plant trees in the grove today. After they are done, there will be 21 trees. How many \\ trees did the grove workers plant today? \\
    Let's think step by step \\ 
    First there are 15 trees. \\ 
    Then there were 21 trees after some more were planted. \\
    So there must have been 21 - 15 = 6 trees. \\
    The answer is 6. \\ 
    \\
    Question: If there are 3 cars in the parking lot and 2 more cars arrive, how many cars are in the parking lot? \\
    Let's think step by step \\
    First there are 3 cars. \\
    Then 2 more cars arrive. \\
    So there must be 3 + 2 = 5 cars. \\
    The answer is 5. \\
    \\ 
    Question: Leah had 32 chocolates and her sister had 42. If they ate 35, how many pieces do they have left in total? \\
    Let's think step by step \\
    First Leah had 32 chocolates and her sister had 42 chocolates. \\
    So in total they had 32 + 42 = 74 chocolates. \\
    Then they ate 35 chocolates. \\
    So there must be 74 - 35 = 39 chocolates. \\
    The answer is 39. \\
    \\
    Question: Jason had 20 lollipops. He gave Denny some lollipops. Now Jason has 12 lollipops. How many lollipops did Jason give to Denny? \\
    Let's think step by step \\
    First Jason had 20 lollipops. \\
    Then he had 12 after giving some to Denny. \\
    So he gave Denny 20 - 12 = 8 lollipops. \\
    The answer is 8. \\
    \\
    Question: Shawn has five toys. For Christmas, he got two toys each from his mom and dad. How many toys does he have now? \\
    Let's think step by step \\
    First Shawn has 5 toys. \\
    Then he got 2 toys each from his mom and dad. \\
    So he must have 5 + 4 = 9 toys. \\
    The answer is 9. \\
    \\
    Question: There were nine computers in the server room. Five more computers were installed each day, from monday to thursday. \\ How many computers are now in the server room? \\
    Let's think step by step \\
    First there were 9 computers. \\
    Then for each of 4 days, 5 more computers were added. \\
    So 5 * 4 = 20 computers were added. \\
    So there must be in total 9 + 20 = 29 computers. \\
    The answer is 29. \\
    \\
    Question: Michael had 58 golf balls. On tuesday, he lost 23 golf balls. On wednesday, he lost 2 more. How many golf balls did he have at the end of Wednesday? \\
    Let's think step by step \\
    First Michael started with 58 golf balls. \\
    Then he lost 23 on Tuesday. \\
    So he had 58 - 23 = 35 golf balls. \\
    Then he lost 2 more on Wednesday. \\
    So he must have 35 - 2 = 33 golf balls. \\
    The answer is 33. \\
    \\
    Question: Olivia has \$23. She bought five bagels for \$3 each. How much money does she have left? \\
    Let's think step by step \\
    First Olivia has 23 dollars. \\
    Then she bought five bagels for 3 dollars each. \\
    We know 5 bagels for 3 dollars each will be 5 * 3 = 15 dollars. \\
    So she has 23 - 15 = 8 dollars left. \\
    The answer is 8. \\
    \hline
  \end{tabular}
  }
  \caption{\label{app-tab: cot-prompt}
     Full prompt for \textbf{CoT} prompting for arithmetic reasoning.
  }
\end{table*}

\begin{table*}
  \centering
  \resizebox{\textwidth}{!}{%
  \begin{tabular}{l}
    \hline

    Question: There are 15 trees in the grove. Grove workers will plant trees in the grove today. After they are done, there will be 21 trees. \\ How many trees did the grove workers plant today? \\
    21 - 15 = 6. \\
The answer is 6. \\
    \\

Question: If there are 3 cars in the parking lot and 2 more cars arrive, how many cars are in the parking lot? \\
3 + 2 = 5. \\
The answer is 5. \\
\\

Question: Leah had 32 chocolates and her sister had 42. If they ate 35, how many pieces do they have left in total? \\
32 + 42 = 74. \\
74 - 35 = 39. \\
The answer is 39. \\
\\
Question: Jason had 20 lollipops. He gave Denny some lollipops. Now Jason has 12 lollipops. How many lollipops did Jason give to Denny? \\
20 - 12 = 8. \\
The answer is 8. \\
\\
Question: Shawn has five toys. For Christmas, he got two toys each from his mom and dad. How many toys does he have now? \\
5 + 4 = 9. \\
The answer is 9. \\
\\
Question: There were nine computers in the server room. Five more computers were installed each day, from monday to thursday. How many \\ computers are now in the server room? \\
5 * 4 = 20. \\
9 + 20 = 29. \\
The answer is 29. \\
\\
Question: Michael had 58 golf balls. On tuesday, he lost 23 golf balls. On wednesday, he lost 2 more. How many golf balls did he have at \\ the end of Wednesday? \\
58 - 23 = 35. \\
35 - 2 = 33. \\
The answer is 33. \\
\\
Question: Olivia has \$23. She bought five bagels for \$3 each. How much money does she have left? \\
5 * 3 = 15. \\
23 - 15 = 8. \\
The answer is 8. \\
    \hline
  \end{tabular}
  }
  \caption{Full prompt for \textbf{w/o text CoT} prompting for arithmetic reasoning.}
    \label{app-tab: without_equation}
\end{table*}

\begin{table*}
  \centering
  \resizebox{\textwidth}{!}{%
  \begin{tabular}{l}
    \hline

    Question: There are 15 trees in the grove. Grove workers will plant trees in the grove today. After they are done, there will be 21 trees. \\ How many trees did the grove workers plant today? \\
Let's think step by step \\
First there are 15 trees. \\
Then there were 21 trees after some more were planted. \\
So there must have been 6 trees. \\
The answer is 6. \\
\\
Question: If there are 3 cars in the parking lot and 2 more cars arrive, how many cars are in the parking lot? \\
Let's think step by step \\
First there are 3 cars. \\
Then 2 more cars arrive. \\
So there must be 5 cars. \\
The answer is 5. \\
\\
Question: Leah had 32 chocolates and her sister had 42. If they ate 35, how many pieces do they have left in total? \\
Let's think step by step \\
First Leah had 32 chocolates and her sister had 42 chocolates. \\
So in total they had 74 chocolates. \\
Then they ate 35 chocolates. \\
So there must be 39 chocolates. \\
The answer is 39. \\
\\
Question: Jason had 20 lollipops. He gave Denny some lollipops. Now Jason has 12 lollipops. How many lollipops did Jason give to Denny? \\
Let's think step by step \\
First Jason had 20 lollipops. \\
Then he had 12 after giving some to Denny. \\
So he gave Denny 8 lollipops. \\
The answer is 8. \\
\\
Question: Shawn has five toys. For Christmas, he got two toys each from his mom and dad. How many toys does he have now? \\
Let's think step by step \\
First Shawn has 5 toys. \\
Then he got 2 toys each from his mom and dad. \\
So he must have 9 toys. \\
The answer is 9. \\
\\
Question: There were nine computers in the server room. Five more computers were installed each day, from monday to thursday. How many \\ computers are now in the server room? \\
Let's think step by step \\
First there were 9 computers. \\
Then for each of 4 days, 5 more computers were added. \\
So 20 computers were added. \\
So there must be in total 29 computers. \\
The answer is 29. \\
\\
Question: Michael had 58 golf balls. On tuesday, he lost 23 golf balls. On wednesday, he lost 2 more. How many golf balls did he have at \\ the end of Wednesday? \\
Let's think step by step \\
First Michael started with 58 golf balls. \\
Then he lost 23 on Tuesday. \\
So he had 35 golf balls. \\
Then he lost 2 more on Wednesday. \\
So he must have 33 golf balls. \\
The answer is 33. \\
\\
Question: Olivia has \$23. She bought five bagels for \$3 each. How much money does she have left? \\
Let's think step by step \\
First Olivia has 23 dollars. \\
Then she bought five bagels for 3 dollars each. \\
We know 5 bagels for 3 dollars each will be 15 dollars. \\
So she has 8 dollars left. \\
The answer is 8. \\
    \hline
  \end{tabular}
  }
  \caption{ Full prompt for \textbf{w/o equation CoT} prompting for arithmetic reasoning.}
\label{app-tab: without-text}
\end{table*}

\begin{table*}
  \centering
  \resizebox{\textwidth}{!}{%
  \begin{tabular}{l}
    \hline

    Question: There are 15 trees in the grove. Grove workers will plant trees in the grove today. After they are done, there will be 21 trees. \\ How many trees did the grove workers plant today? \\
Let's think step by step \\
First there are 15 trees. \\
Then there were 21 trees after some more were planted. \\
So there must have been 21 - 15 = 1 trees. \\
The answer is 1. \\
\\
Question: If there are 3 cars in the parking lot and 2 more cars arrive, how many cars are in the parking lot? \\
Let's think step by step \\
First there are 3 cars. \\
Then 2 more cars arrive. \\
So there must be 3 + 2 = 3 cars. \\
The answer is 3. \\
\\
Question: Leah had 32 chocolates and her sister had 42. If they ate 35, how many pieces do they have left in total? \\
Let's think step by step \\
First Leah had 32 chocolates and her sister had 42 chocolates. \\
So in total they had 32 + 42 = 12 chocolates. \\
Then they ate 35 chocolates. \\
So there must be 12 - 35 = 45 chocolates. \\
The answer is 45. \\
\\
Question: Jason had 20 lollipops. He gave Denny some lollipops. Now Jason has 12 lollipops. How many lollipops did Jason give to Denny? \\
Let's think step by step \\
First Jason had 20 lollipops. \\
Then he had 12 after giving some to Denny. \\
So he gave Denny 20 - 12 = 2 lollipops. \\
The answer is 2. \\
\\
Question: Shawn has five toys. For Christmas, he got two toys each from his mom and dad. How many toys does he have now? \\
Let's think step by step \\
First Shawn has 5 toys. \\
Then he got 2 toys each from his mom and dad. \\
So he must have 5 + 4 = 6 toys. \\
The answer is 6. \\
\\
Question: There were nine computers in the server room. Five more computers were installed each day, from monday to thursday. How many \\ computers are now in the server room? \\
Let's think step by step \\
First there were 9 computers. \\
Then for each of 4 days, 5 more computers were added. \\
So 5 * 4 = 22 computers were added. \\
So there must be in total 9 + 22 = 23 computers. \\
The answer is 22. \\
\\
Question: Michael had 58 golf balls. On tuesday, he lost 23 golf balls. On wednesday, he lost 2 more. How many golf balls did he have at \\ the end of Wednesday? \\
Let's think step by step \\
First Michael started with 58 golf balls. \\
Then he lost 23 on Tuesday. \\
So he had 58 - 23 = 14 golf balls. \\
Then he lost 2 more on Wednesday. \\
So he must have 14 - 2 = 23 golf balls. \\
The answer is 23. \\

Question: Olivia has \$23. She bought five bagels for \$3 each. How much money does she have left? \\
Let's think step by step \\
First Olivia has 23 dollars. \\
Then she bought five bagels for 3 dollars each. \\
We know 5 bagels for 3 dollars each will be 5 * 3 = 23 dollars. \\
So she has 23 - 23 = 9 dollars left. \\
The answer is 9. \\
    \hline
  \end{tabular}
  }
  \caption{ Full prompt for \textbf{Incorrect CoT} prompting for our RQ3 experiment.}
\label{app-tab: incorrect-cot}
\end{table*}

\begin{table*}
  \centering
  \resizebox{\textwidth}{!}{%
  \begin{tabular}{l}
    \hline

    Question: There are 15 trees in the grove. Grove workers will plant trees in the grove today. After they are done, there will be 21 trees. \\ How many trees did the grove workers plant today? \\
Let's think step by step \\
First there are 15 trees. \\
Then there were 21 trees after some more were planted. \\
So there must have been 21 - 15 = Dawson trees. \\
The answer is Dawson. \\
\\
Question: If there are 3 cars in the parking lot and 2 more cars arrive, how many cars are in the parking lot? \\
Let's think step by step \\
First there are 3 cars. \\
Then 2 more cars arrive. \\
So there must be 3 + 2 = Russel cars. \\
The answer is Russel. \\
\\
Question: Leah had 32 chocolates and her sister had 42. If they ate 35, how many pieces do they have left in total? \\
Let's think step by step \\
First Leah had 32 chocolates and her sister had 42 chocolates. \\
So in total they had 32 + 42 = Cedric chocolates. \\
Then they ate 35 chocolates. \\
So there must be Cedric - 35 = Hendricks chocolates. \\
The answer is Hendricks. \\
\\
Question: Jason had 20 lollipops. He gave Denny some lollipops. Now Jason has 12 lollipops. How many lollipops did Jason give to Denny? \\
Let's think step by step \\
First Jason had 20 lollipops. \\
Then he had 12 after giving some to Denny. \\
So he gave Denny 20 - 12 = Edison lollipops. \\
The answer is Edison. \\
\\
Question: Shawn has five toys. For Christmas, he got two toys each from his mom and dad. How many toys does he have now? \\
Let's think step by step \\
First Shawn has 5 toys. \\
Then he got 2 toys each from his mom and dad. \\
So he must have 5 + 4 = Kaylin toys. \\
The answer is Kaylin. \\
\\
Question: There were nine computers in the server room. Five more computers were installed each day, from monday to thursday. How many \\ computers are now in the server room? \\\
Let's think step by step \\
First there were 9 computers. \\
Then for each of 4 days, 5 more computers were added. \\
So 5 * 4 = Mcgowan computers were added. \\
So there must be in total 9 + Mcgowan = Damarion computers. \\
The answer is Damarion. \\
\\
Question: Michael had 58 golf balls. On tuesday, he lost 23 golf balls. On wednesday, he lost 2 more. How many golf balls did he have at \\ the end of Wednesday? \\
Let's think step by step \\
First Michael started with 58 golf balls. \\
Then he lost 23 on Tuesday. \\
So he had 58 - 23 = Jemima golf balls. \\
Then he lost 2 more on Wednesday. \\
So he must have Jemima - 2 = Astrid golf balls. \\
The answer is Astrid. \\
\\
Question: Olivia has \$23. She bought five bagels for \$3 each. How much money does she have left? \\
Let's think step by step \\
First Olivia has 23 dollars. \\
Then she bought five bagels for 3 dollars each. \\
We know 5 bagels for 3 dollars each will be 5 * 3 = Gallagher dollars. \\
So she has 23 - Gallagher = Baily dollars left. \\
The answer is Baily. \\
    \hline
  \end{tabular}
  }
\caption{ Full prompt for Out-of-distribution  \textbf{(OOD) CoT} prompting for arithmetic reasoning.}
\label{app-tab: ood_cot}
\end{table*}

\begin{table*}
  \centering
  \resizebox{\textwidth}{!}{%
  \begin{tabular}{l}
    \hline

    Question: If there are 3 cars in the parking lot and 2 more cars arrive, how many cars are in the parking lot? \\
Let's think step by step \\
First there are 3 cars. \\
Then 2 more cars arrive. \\
So there must be 3 + 2 = 5 cars. \\
The answer is 5. \\
\\
Question: Paddington has 40 more goats than Washington. If Washington has 140 goats, how many goats do they have in total? \\
Let's think step by step \\
First Paddington has 40 more goats than Washington. \\
We know Washington has 140 goats. \\
So Paddington has 40 + 140 = 180 goats. \\
So they have 180 + 140 = 320 goats in total. \\
The answer is 320. \\
\\
Question: Christina has 3 snakes. 1 snake is 2 feet long. Another snake is 16 inches long. The last snake is 10 inches long. How many \\ inches are all of her snakes combined? \\
Let's think step by step \\
First Christina has 3 snakes. \\
Then 1 snake is 2 feet long. \\
We know 1 foot is 12 inches. \\
So 2 feet is 12 + 12 = 24 inches. \\
Then another snake is 16 inches long. \\
Then the last snake is 10 inches long. \\
So all of her snakes combined are 24 + 16 + 10 = 50 inches. \\
The answer is 50. \\
\\
Question: Bush and Matt are brothers. Bush is younger than Matt by 3 years. This year Bush will be 12 years old. What will be Matt's \\ age 10 years from now? \\
Let's think step by step \\
First Bush is younger than Matt by 3 years. \\
We know Bush will be 12 years old this year. \\
So Matt will be 12 + 3 = 15 years old this year. \\
Then Matt's age 10 years from now will be 15 + 10 = 25 years old. \\
The answer is 25. \\
\\
Question: Jeremy listened to five more songs yesterday than today. Yesterday, he listened to nine songs. How many songs did Jeremy listen \\ to in two days? \\
Let's think step by step \\
First Jeremy listened to 9 songs yesterday. \\
Then he listened to 5 more songs yesterday than today. \\
So he listened to 9 + 5 = 14 songs today. \\
So he listened to 9 + 14 = 23 songs in two days. \\
The answer is 23. \\
\\
Question: Jar A has 28 marbles. Jar B has 12 more marbles than jar A. Jar C has as many marbles as jar B. How many marbles are there altogether? \\
Let's think step by step \\
First Jar A has 28 marbles. \\
Then Jar B has 12 more marbles than jar A. \\
So Jar B has 28 + 12 = 40 marbles. \\
Then Jar C has as many marbles as jar B. \\
So Jar C has 40 marbles. \\
So there are 28 + 40 + 40 = 108 marbles altogether. \\
The answer is 108. \\
\\
Question: Marion received 20 more turtles than Mia at the animal rescue center. If Mia received 40 turtles, how many turtles \\ did they receive together? \\
Let's think step by step \\
First Marion recieved 20 more turtles than Mia. \\
We know Mia received 40 turtles. \\
So Marion received 20 + 40 = 60 turtles. \\
So together they received 60 + 40 = 100 turtles. \\
The answer is 100. \\
\\
Question: Shawn has five toys. For Christmas, he got two toys each from his mom and dad. How many toys does he have now? \\
Let's think step by step \\
First Shawn has 5 toys. \\
Then he got 2 toys each from his mom and dad. \\
So he must have 5 + 4 = 9 toys. \\
The answer is 9. \\
    \hline
  \end{tabular}
  }
\caption{ Full prompt for \textbf{AddOnly CoT} prompting for arithmetic reasoning.}
\label{app-tab: add-only}
\end{table*}

\begin{table*}
  \centering
  \resizebox{\textwidth}{!}{%
  \begin{tabular}{l}
    \hline

    Question: Super Clean Car Wash Company cleans 80 cars per day. They make \$5 per car washed. How much money will they make in 5 days? \\
Let's think step by step \\
First Super Clean Car Wash Company cleans 80 cars per day. \\
Then they make 5 dollars per car washed. \\
So they make 80 x 5 = 400 dollars per day. \\
So they make 400 x 5 = 2000 dollars in 5 days. \\
The answer is 2000. \\
\\
Question: A farmer has 5 cows. Each cow has 4 legs. How many legs do the cows have in total? \\
Let's think step by step \\
First a farmer has 5 cows. \\
Then each cow has 4 legs. \\
So the cows have 5 x 4 = 20 legs in total. \\
The answer is 20. \\
\\
Question: Sam watches two movies each day. Each movie is 2 hours long. How many minutes does Sam spend watching movies in 5 days? \\
Let's think step by step \\
First Sam watches two movies each day. \\
Then each movie is 2 hours long. \\
We know 1 hour is 60 minutes. \\
So 2 hours is 60 x 2 = 120 minutes. \\
So Sam spends 120 x 2 = 240 minutes watching movies each day. \\
So Sam spends 240 x 5 = 1200 minutes watching movies in 5 days. \\
The answer is 1200. \\
\\
Question: Carla has 3 bags. Each bag has 5 apples. How many apples does Carla have in total? \\
Let's think step by step \\
First Carla has 3 bags. \\
Then each bag has 5 apples. \\
So Carla has 3 x 5 = 15 apples in total. \\
The answer is 15. \\
\\
Question: James takes 20 units per semester at community college.  If each unit costs \$50 how much does he pay for 2 semesters? \\
Let's think step by step \\
First James takes 20 units per semester at community college. \\
Then each unit costs 50 dollars. \\
So he pays 20 x 50 = 1000 dollars per semester. \\
So he pays 1000 x 2 = 2000 dollars for 2 semesters. \\
The answer is 2000. \\
\\
Question: In a jar that has 50 ants, the number of ants in the jar doubles each hour. How many ants will be in the jar after 5 hours? \\
Let's think step by step \\
First there are 50 ants in the jar. \\
Then the number of ants in the jar doubles each hour. \\
So there will be 50 x 2 = 100 ants in the jar after 1 hour. \\
So there will be 100 x 2 = 200 ants in the jar after 2 hours. \\
So there will be 200 x 2 = 400 ants in the jar after 3 hours. \\
So there will be 400 x 2 = 800 ants in the jar after 4 hours. \\
So there will be 800 x 2 = 1600 ants in the jar after 5 hours. \\
The answer is 1600. \\
\\
Question: Mark loves to see shows in theaters. He decided to visit the theater at least once a week. One performance lasts 3 hours. The price \\ of the ticket depends on the time spent in the theater and stands at \$5 for each hour. How much will Mark \\ spend on visits to the theater in 6 weeks? \\
Let's think step by step \\
First Mark decided to visit the theater at least once a week. \\
Then one performance lasts 3 hours. \\
We know the price of the ticket depends on the time spent in the theater and stands at 5 dollars for each hour. \\
So the price of the ticket for one performance is 5 x 3 = 15 dollars. \\
So Mark will spend 15 x 6 = 90 dollars on visits to the theater in 6 weeks. \\
The answer is 90. \\
\\
Question: A sixty bulb watt uses 60 watts of power each day. If Allyn has 40 such bulbs in his house and pays an electricity bill of twenty \\ cents per power watt used, calculate Allyn's total monthly expenses on electricity in June. \\
Let's think step by step \\
First a sixty bulb watt uses 60 watts of power each day. \\
Then Allyn has 40 such bulbs in his house. \\
So Allyn has 40 x 60 = 2400 watts of power each day. \\
Then Allyn pays an electricity bill of twenty cents per power watt used. \\
So Allyn pays 2400 x 0.2 = 480 dollars per day. \\
So Allyn pays 480 x 30 = 14400 dollars per month. \\
The answer is 14400. \\
    \hline
  \end{tabular}
  }
\caption{ Full prompt for \textbf{MultOnly CoT} prompting for arithmetic reasoning.}
\label{app-tab: mult-only}
\end{table*}

\end{document}